\documentclass{article}

\PassOptionsToPackage{numbers, sort&compress}{natbib}

\usepackage[preprint]{neurips_2021}

\usepackage[utf8]{inputenc} %
\usepackage[T1]{fontenc}    %
\usepackage{hyperref}       %
\usepackage{url}            %
\usepackage{booktabs}       %
\usepackage{amsfonts}       %
\usepackage{nicefrac}       %
\usepackage{microtype}      %
\usepackage{amssymb}
\usepackage{amsmath}
\usepackage{xcolor}
\usepackage{cancel}

\newcommand{\figleft}{{\em (Left)}}

\newcommand{\figright}{{\em (Right)}}
\newcommand{\figtop}{{\em (Top)}}
\newcommand{\figbottom}{{\em (Bottom)}}

\def\eqref#1{equation~\ref{#1}}

\def\1{\bm{1}}

\DeclareMathAlphabet{\mathsfit}{\encodingdefault}{\sfdefault}{m}{sl}
\SetMathAlphabet{\mathsfit}{bold}{\encodingdefault}{\sfdefault}{bx}{n}

\def\gB{{\mathcal{B}}}

\def\gD{{\mathcal{D}}}

\def\gL{{\mathcal{L}}}
\def\gM{{\mathcal{M}}}

\def\gZ{{\mathcal{Z}}}

\newcommand{\E}{\mathbb{E}}

\usepackage{amsthm}
\usepackage{graphicx}
\usepackage{subcaption}
\usepackage{wrapfig}
\usepackage{amssymb}%
\usepackage{pifont}%
\usepackage{algorithm}
\usepackage[noend]{algpseudocode}
\usepackage[frozencache=true,cachedir=minted-cache]{minted}
\newcommand{\cmark}{\ding{51}}%
\newcommand{\xmark}{\ding{55}}%
\newcommand{\cs}{\mathbf{s_t}}
\newcommand{\ns}{\mathbf{s_{t+1}}}
\newcommand{\ca}{\mathbf{a_t}}
\newcommand{\na}{\mathbf{a_{t+1}}}
\newcommand{\cz}{\mathbf{z_t}}
\newcommand{\nz}{\mathbf{z_{t+1}}}

\newtheorem{theorem}{Theorem}[section]

\newtheorem{lemma}[theorem]{Lemma}

\title{Robust Predictable Control}

\author{%
  Benjamin Eysenbach$^{1 \; 2}$ \\
  \And
  Ruslan Salakhutdinov$^{1}$ \\
  \And
  Sergey Levine$^{2 \; 3}$ \\
  \And
  \vspace{-2.5em} \\
  $^{1}$Carnegie Mellon University, \quad $^{2}$Google Brain, \quad $^{3}$UC Berkeley \\
  \texttt{beysenba@cs.cmu.edu}
}

\begin{document}

\maketitle

\begin{abstract}
Many of the challenges facing today's reinforcement learning (RL) algorithms, such as robustness, generalization, transfer, and computational efficiency are closely related to compression.  Prior work has convincingly argued why minimizing information is useful in the supervised learning setting, but standard RL algorithms lack an explicit mechanism for compression. The RL setting is unique because (1) its sequential nature allows an agent to use past information to avoid looking at future observations and (2) the agent can optimize its behavior to prefer states where decision making requires few bits. We take advantage of these properties to propose a method (RPC) for learning \emph{simple} policies. This method brings together ideas from information bottlenecks, model-based RL, and bits-back coding into a simple and theoretically-justified algorithm. Our method jointly optimizes a latent-space model and policy to be \emph{self-consistent}, such that the policy avoids states where the model is inaccurate. We demonstrate that our method achieves much tighter compression than prior methods, achieving up to 5$\times$ higher reward than a standard information bottleneck. We also demonstrate that our method learns policies that are more robust and generalize better to new tasks.\footnote{Project site with videos and code: \url{https://ben-eysenbach.github.io/rpc}}
\end{abstract}

\vspace{-0.7em}
\section{Introduction}
\vspace{-0.7em}

Many areas of reinforcement learning (RL) research focus on specialized problems, such as learning invariant representations, improving robustness to adversarial attacks, improving generalization, or building better world models. These problems are often symptoms of a deeper underlying problem: autonomous agents use too many bits from their environment. For the purpose of decision making, most information about the world is irrelevant. For example, a lane keeping feature on a car may take as input high-resolution camera input (millions of bits), but only needs to extract a few bits of information about the relative orientation of the car in the lane. Agents that rely on more bits of information run the risk of overfitting to the training task.

Agents that use few bits of information gain a number of appealing properties. 
These agents can better cope with high-dimensional sensory inputs (e.g., dozens of cameras on a self-driving car) and will be forced to learn representations that are more broadly applicable. Agents that throw away most information will be agnostic to idiosyncrasies in observations, providing robustness to missing or corrupted observations and better transfer to different scenarios. For example, if an agent ignores 99.9\% of bits, then corrupting a random bit is unlikely to change the agent's behavior. Moreover, an agent that minimizes bits will prefer states where the dynamics are easy to predict, meaning that the agent's resulting behavior will be easier to model.
Thus, compression not only changes an agent's representation, but also changes its behavior: an agent that can only use a limited number of bits will avoid risky behaviors that require more bits to execute (see Fig.~\ref{fig:two_way_qualitative}).

The generalization and robustness of a machine learning model is directly related to the complexity of that model. Indeed, standard techniques for reducing complexity, such as the information bottleneck~\citep{tishby2015deep, achille2018emergence}, can be directly applied to the RL setting~\citep{goyal2019infobot, igl2019generalization, lu2020dynamics}. While these approaches make the policy's action a simple function of the state, they ignore the temporal dimension of decision making. Instead, we will focus on learning policies whose temporally-extended behavior is simple. Our key observation is that \emph{a policy's behavior is simple if it is predictable}. Throughout this paper, we will use compression as a measure of model complexity.

Our method improves upon prior methods that apply an information bottleneck to RL~\citep{goyal2019infobot, igl2019generalization, lu2020dynamics} by recognizing two important properties of the decision making setting. 
\textbf{First}, because agents make a sequence of decisions, they can use salient information at one time step to predict salient information at the next time step. These predictions can decrease the amount of information that the agent needs to sense from the environment. We will show that learning a predictive model is not an ad-hoc heuristic, but rather a direct consequence of minimizing information using bits-back coding~\citep{hinton1993keeping, frey1997efficient}. \textbf{Second}, unlike supervised learning, the agent can change the distribution over states, choosing behaviors that visit states that are easier to compress.
For example, imagine driving on a crowded road. Aggressively passing and tailgating other cars may result in reaching the destination faster, but requires careful attention to other vehicles and fast reactions.
In contrast, a policy optimized for using few bits would not pass other cars and would leave a larger following distance (see Fig.~\ref{fig:acc_qualitative}).
Combined, these two capabilities result in a method that jointly trains a latent space model and a control policy, with the policy being rewarded for visiting states where that model is accurate. Unlike typical model-based methods, our method explicitly optimizes for the accuracy of open-loop planning and results in a model and policy that are \emph{self-consistent}.

The main contribution of this paper is an RL algorithm, \textbf{robust predictable control (RPC)}, for learning policies that use few bits of information. We will refer to such policies as \emph{compressed policies}.
RPC brings together ideas from information bottlenecks, model-based RL, and bits-back coding into a simple and theoretically-justified algorithm. RPC achieves much higher compression than prior methods; for example, RPC achieves $\sim\!5\times$ higher return than a standard information bottleneck compared at the same bitrate.
Experiments demonstrate practical benefits from the compressed policies learned by RPC: they are more robust than those learned by alternative approaches, generalize well to new tasks, and learn behaviors that can be composed for hierarchical RL.

\vspace{-0.7em}
\section{Related Work}
\vspace{-0.7em}

The problem of learning simple models has a long history in the machine learning community~\citep{lecun1990optimal, hinton1993keeping}.
Simplicity is often measured by the mutual information between its inputs and outputs~\citep{bassily2018learners, tishby2015deep}, a metric that has been used to study the generalization properties~\citep{tishby2015deep} and representations learned by~\citep{achille2018emergence} neural networks. Our work extends these results to the RL setting by observing that the agent can change its behavior (i.e., the data distribution) to be more easily compressed.
In the RL community, prior work has used the variational information bottleneck (VIB)~\citep{alemi2016deep} to minimize communication between agents in a multi-agent setting~\citep{wang2020learning} and to improve exploration in a goal-reaching setting~\citep{goyal2019infobot}.
The most related RL methods are those that use an information bottleneck in RL to improve generalization~\citep{igl2019generalization, lu2020dynamics}.
Whereas these prior methods compress observations individually, we will compress \emph{sequences} of observations. This difference, which corresponds to learning a latent-space model, improves compression and increases robustness on downstream tasks.

In the RL setting, compression offers a tool for studying problems such as representation learning, robustness and generalization.
Prior RL methods have used mutual information to learn representations that are good for control~\citep{oord2018representation, laskin2020curl, gelada2019deepmdp, lee2019stochastic}. Our method learns a representation of observations that, like contrastive learning methods~\citep{oord2018representation, nachum2018near, laskin2020curl}, avoids the need to reconstruct the observation.
While these contrastive methods maximize the mutual information between representations across time, our method will minimize the mutual information between observations and representations. Moreover, we will jointly optimize the policy and representation using a single, unified objective.
Our method is also related to robust RL, which studies the problem of learning RL policies that are resilient against perturbations to the environment~\citep{morimoto2005robust, huang2017adversarial, tessler2019action, kamalaruban2020robust}. While prior robust RL methods typically involve solving a two-player game, we show that compression is a simpler mechanism for achieving some robustness benefits.
Finally, the problem of learning RL policies that generalize has been studied by many prior papers~\citep{cobbe2019quantifying, farebrother2018generalization}. We will show that compression also yields RL policies that generalize well.

\vspace{-0.7em}
\section{Reinforcement Learning with Fewer Bits}
\vspace{-0.7em}

This section introduces the idea that predicting the future allows RL policies to operate with fewer bits. We derive this idea from first principles, develop it into a complete RL method, then discuss connections with model-based RL, bits-back coding, and other related topics.

\clearpage
\subsection{Notation and Preliminaries}
\label{sec:notation}

An agent interacts in an MDP defined by states $\cs$ and actions $\ca$.\footnote{We use the terms ``states'' and ``observations'' interchangeably.}
The agent samples actions from a policy $\pi_\theta(\ca \mid \cs, \mathbf{s_{t-1}}, \mathbf{s_{t-2}}, \cdots)$.\footnote{We introduce the policy as being conditioned on states because, even if the task is Markovian, the agent's choice to omit certain bits from the representation can turn an otherwise fully-observed task into a partially-observed task.}
We will construct this policy by learning an encoder $\phi(\cz \mid \cs)$ (which produces a representation $\cz$ of the current state $\cs$), and a high-level policy $\pi_\theta^z(\ca \mid \cz)$ which \emph{decodes} a representation $\cz$ into an action $\ca$.
The environment dynamics are defined by an initial state $\mathbf{s_1} \sim p_1(\mathbf{s_1})$ and a transition function $p(\ns \mid \cs, \ca)$.
The standard RL objective is to maximize the expected $\gamma$-discounted sum of rewards $r(\cs, \ca)$: $\max_\pi \E_\pi\left[\sum_{t=1}^\infty \gamma^t r(\cs, \ca) \right]$.
The discount factor can be interpreted as saying that the episode terminates at every time step with probability $(1 - \gamma)$, an interpretation we will use in Sec.~\ref{sec:our-objective}.

A model is simpler if it expresses a simpler input-output relationship~\citep{tishby2015deep, bassily2018learners}.
We will measure the complexity of a function using an \emph{information bottleneck}, which is the mutual information $I(\mathbf{x}; \mathbf{y})$ between an input $\mathbf{x}$ and an output $\mathbf{y}$~\citep{tishby2015deep, achille2018emergence}. Mutual information is closely tied to the energy required to implement that function on an ideal physical system~\citep{ortega2013thermodynamics}.
The \emph{variational information bottleneck} provides a tractable upper bound on mutual information~\citep{alemi2016deep, achille2018emergence}:
\begin{equation*}
    I(\mathbf{x}; \mathbf{y}) \le \E_{p(\mathbf{x}, \mathbf{y})}\left[\log \left( \frac{p(\mathbf{y} \mid \mathbf{x})}{m(\mathbf{y})}\right) \right],
\end{equation*}
where $m(\mathbf{y})$ is an arbitrary prior distribution. Applying the information bottleneck to an intermediate layer $\mathbf{z} = \phi(\mathbf{x})$ is sufficient for bounding the mutual information between input $x$ and output~$y$.  %

Following prior work on compression in RL~\citep{igl2019generalization, lu2020dynamics}, we aim to maximize rewards while minimizing the number of bits. While there are many ways to apply compression to RL (e.g., compressing actions, goals, or individual observations), we will focus on compressing \emph{sequences} of states.
The input is a sequence of states, $\mathbf{s_{1:\infty}} \triangleq (\mathbf{s_1}, \mathbf{s_2}, \cdots)$; the output is a sequence of actions, $\mathbf{a_{1:\infty}} \triangleq (\mathbf{a_1}, \mathbf{a_2}, \cdots)$. The objective is to learn a representation $\phi_\theta(\cz \mid \cs)$ and policy $\pi_\theta(\ca \mid \cz)$ that maximize reward, subject to the constraint that the policy uses on average $C > 0$ bits of information per episode:
\begin{equation}
    \max_{\theta} \E_{\pi, \phi}\left[\sum_{t=1}^\infty \gamma^t r(\cs, \ca) \right] \qquad \text{s.t.} \quad \E_{\pi}[I(\mathbf{s_{1:\infty}}; \mathbf{z_{1:\infty}})] \le C, \label{eq:main-obj}
\end{equation}
While prior work on compression in RL~\citep{igl2019generalization, lu2020dynamics} has applied the VIB to states \emph{independently}, we will aim to compress entire \emph{sequences} of observations.
Applying the VIB to sequences allows us to use more expressive choices of the prior $m(\mathbf{z_{1:\infty}})$. Our prior will
use previous representations to \emph{predict} the prior for the next representation, allowing us to obtain a tighter bound on mutual~information.

\subsection{Using Fewer Bits by Predicting the Future}
\label{sec:our-objective}

The main idea of our method is that if the agent can accurately predict the future, then the agent will not need to observe as many bits from future observations. Precisely, the agent will learn a latent dynamics model that predicts the next representation using the current representation and action. In addition to predicting the future, the agent can also decrease the number of bits by changing its behavior. States where the dynamics are hard to predict will require more bits, so the agent will prefer visiting states where its learned model can accurately predict the next state.

This intuition corresponds exactly to solving the optimization problem in Eq.~\ref{eq:main-obj} with a prior that is factored autoregressively: $m_{1:\infty}(\mathbf{z_{1:\infty}} \mid \mathbf{a_{1:\infty}}) = m(\mathbf{z_1}) \prod_t m_\theta(\nz \mid \cz, \ca)$. Note that the prior has learnable parameters $\theta$.
We apply the VIB to obtain an upper bound on the constraint in Eq.~\ref{eq:main-obj}:
{\footnotesize 
\begin{equation}
    \E_{\pi}[I(\mathbf{s_{1:\infty}}; \mathbf{z_{1:\infty}})] \le \E_{\pi} \left[\log \left(\frac{p(\mathbf{z_{1:\infty}} \mid \mathbf{s_{1:\infty}})}{m(\mathbf{z_{1:\infty}})} \right) \right] = \E_{\pi}\left[ \sum_{t=1}^\infty \gamma^t (\log \phi_\theta(\cz \mid \cs) - \log m_\theta(\nz \mid \cz, \ca)) \right]. \label{eq:info-cost}
\end{equation}}
This objective is different from prior work that applies the VIB to RL~\citep{goyal2019infobot, igl2019generalization, lu2020dynamics} because the prior $m_\theta(\cz)$ is \emph{predicted}, rather than fixed to be a unit Normal distribution.
The discount factor $\gamma$ reflects the assumption from Sec.~\ref{sec:notation} that the episode terminates with probability $(1 - \gamma)$ at each time step; of course, no bits are used after the episode terminates. Our final objective optimizes the policy $\pi$, the encoder $\phi$, and the prior $m$ to maximize reward and minimizing information:
{\footnotesize \begin{equation}
    \max_{\theta} \E_{\pi_\theta, \phi_\theta, m_\theta}\left[\sum_{t=1}^\infty \gamma^t r(\cs, \ca) \right] \qquad \text{s.t.} \quad \E_{\pi_\theta, \phi_\theta, m_\theta} \left[\sum_t \gamma^t \left(\log \phi_\theta(\cz \mid \cs) - \log m_\theta(\nz \mid \cz, \ca) \right) \right] \le C. \label{eq:objective}
\end{equation}}
For the encoder, this objective looks like a modification of the information bottleneck where the prior is \emph{predicted} from the previous representation and action.
Our compression objective modifies the MDP by imposing an \emph{information cost} associated with gathering information. This new MDP has a reward~function
\begin{equation}
\tilde{r}_\lambda(\cs, \ca, \ns) \triangleq r(\cs, \ca) + \lambda \underbrace{\left(\log m_\theta(\nz \mid \cz, \ca) - \log \phi_\theta(\nz \mid \ns) \right)}_{\text{(negative of the) information cost}}, \label{eq:reward-and-info}
\end{equation}
where $\lambda$ is the cost per bit. The term $\log \phi_\theta(\cz \mid \cs)$ corresponds to the number of bits required to represent the representation $\cz$. The term $\log m_\theta(\nz \mid \cz, \ca)$ reflects how well the agent can predict the next representation. In effect, the agent has to ``pay'' for bits of the observation, but it gets a ``refund'' on bits that it predicted from the previous time step, analogous to bits-back coding~\citep{frey1997efficient, hinton1993keeping}.
Said in other words, we want to minimize the \emph{excess} bits required to infer $\cz$ from the current state, relative to the number of bits required to predict $\cz$ from the previous ($z_{t-1}, a_{t-1}$).
Note that the standard VIB approaches to RL ~\citep{goyal2019infobot, igl2019generalization, lu2020dynamics} do not receive this refund because they compress observations independently, rather than compressing \emph{sequences} of observations.
Importantly, the agent optimizes not only its representation but also its behavior to minimize this information cost: the agent learns a representation that is easily predictable and learns to visit states where that representation is easily predictable. We therefore call our method \textbf{robust predictable control (RPC)}.

\section{A Practical Algorithm}
\label{sec:alg}

\begin{wrapfigure}[15]{R}{0.5\textwidth}
    \vspace{-1.5em}
    \centering
    \includegraphics[width=\linewidth]{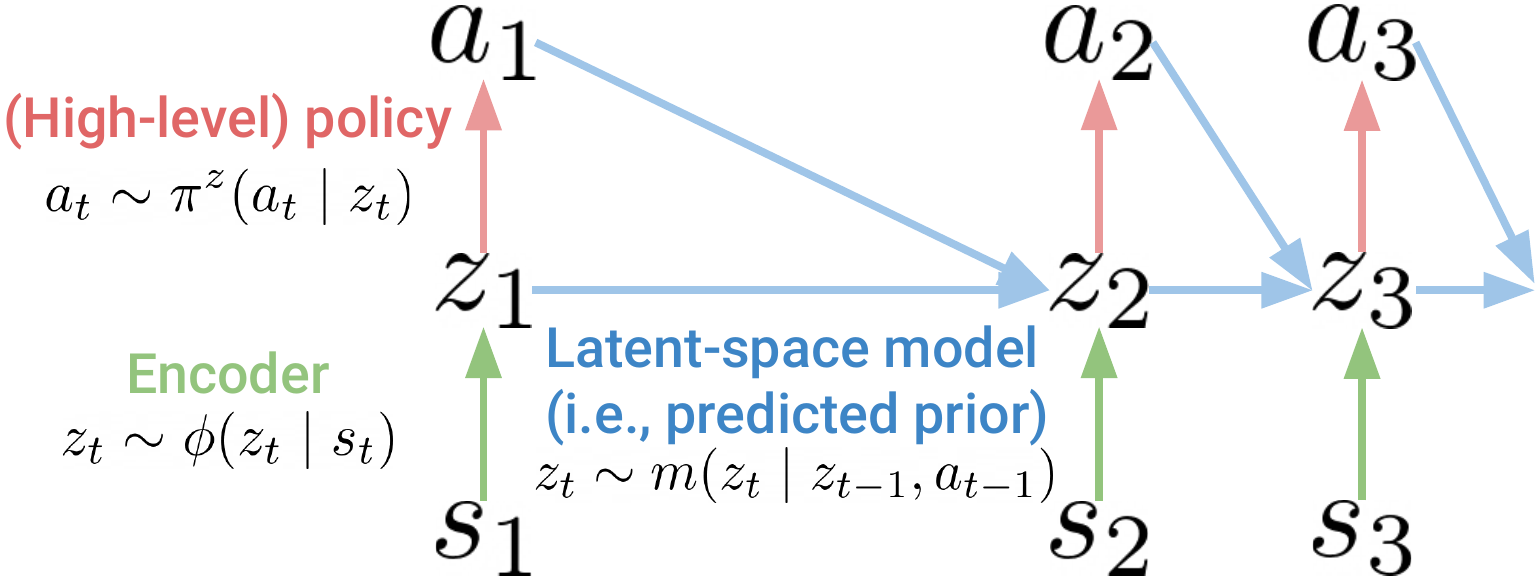}
    \caption{{\footnotesize \textbf{Robust Predictable Control (RPC)}: Our method learns three components: an encoder $\phi(\cz \mid \cs)$, a latent-space model $m(\cz \mid \mathbf{z_{t-1}}, \mathbf{a_{t-1}})$, and a policy $\pi^z(\ca \mid \cz)$.
    All three are trained to be self-consistent: the representation of the next state should be equal to the representation predicted by the model.
    In contrast, a conventional VIB~\citep{igl2019generalization, lu2020dynamics} omits the {\color{blue} blue arrows}.
    }}
    \label{fig:pgm}
\end{wrapfigure}
Our method for optimizing (Eq.~\ref{eq:objective}) is an actor-critic method applied to the information-augmented reward function in Eq.~\ref{eq:reward-and-info}.
We introduce a Q function $Q_\psi(\cs, \ca) = \E\left[\sum_t \gamma^t \tilde{r}(\cs, \ca, \ns) \right]$ for estimating the expected future returns of information-regularized reward function $\tilde{r}$. We optimize this Q function using standard temporal difference learning:
\begin{equation}
    \gL(\psi) = \frac{1}{2}(Q_\psi(\cs, \ca) - \mathbf{y_t})^2 \label{eq:q-loss}
\end{equation}
where $\mathbf{y_t} = \mathbf{\tilde{r}_t} + \gamma Q_\psi(\ns, \na)$.
In practice, we use a separate target Q function and do not backpropagate gradients through the target Q function.
Since our overall objective (Eq.~\ref{eq:main-obj}) only entails compressing the policy, not the Q function, we can condition the Q function directly on the state. In effect, we provide the Q-function with ``side'' information that is not available to the policy, similarly to prior work~\citep{levine2016end, pinto2017asymmetric, akkaya2019solving}.

To optimize the encoder, prior, and policy, we note that our overall objective (Eq.~\ref{eq:main-obj}) can be expressed in terms of the immediate reward plus the Q function (see derivation in Appendix~\ref{appendix:one-step-obj}):
\begin{equation}
    \gL(\theta; \cs) = \E_{\substack{\mathbf{z_{t-1}} \sim \phi_\theta(\mathbf{s_{t-1}}), \cs \sim p(\cs \mid \mathbf{s_{t-1}}, \mathbf{a_{t-1}}) \\ \cz \sim \phi(\cs), \ca \sim \pi_\theta^z(\ca \mid \cz)}}[Q_\psi(\cs, \ca) +  \lambda \left(\log m_\theta(\cz \mid \mathbf{z_{t-1}}, \mathbf{a_{t-1}}) - \log \phi_\theta(\cz \mid \cs) \right)]. \label{eq:one-step-obj}
\end{equation}
Since the encoder is stochastic, we compute gradients through it using the reparametrization trick.
The fact that all three components are optimized with respect to the same objective makes implementation of RPC surprisingly simple.
Note that RPC does not require sampling from $m_\theta(\nz \mid \cz, \ca)$ and does not require backpropagation through time.
In our implementation, we instantiate the encoder $\phi_\theta(\cs)$, the prior $m_\theta(\nz \mid \cz, \ca)$, and the high-level policy $\pi_\theta^z(\ca \mid \cz)$ as neural networks with parameters $\theta$. The Q function $Q_\psi(\cs, \ca)$ is likewise represented as a neural network with parameters $\phi$.
We update the dual parameter $\lambda \ge 0$ using dual gradient descent.
We use standard tricks such as target networks and taking the minimum over two Q values. We refer the reader to Appendix~\ref{appendix:details} and the open-sourced code for details.

\section{Connections and Analysis}

In this section we discuss the connections between robust predictable control and other areas of RL. We then prove that RPC enjoys certain robustness guarantees.

\subsection{Connections}
\label{sec:connections}

RPC is closely related to a number of ideas in the RL literature. This section explains these connections, with the aim of building intuition into how RPC works and providing an explanation for why RPC should learn robust policies and useful representations.
We include a further discussion of the relationship to MaxEnt RL, the expression for the optimal encoder, and the value of information in Appendix~\ref{appendix:optimizing-encoder-maxent}.

\paragraph{Model-Based RL.}

The prior $m_\theta(\cz \mid \mathbf{z_{t-1}}, \mathbf{a_{t-1}})$ learned by RPC can be viewed as a dynamics model. Rather than predicting what the next state will be, this model predicts what the representation of the next state will be.  While the model is trained to make accurate predictions, \emph{the policy is also trained to visit states and actions where the model will be more accurate.}

\paragraph{Representation learning.}
\label{sec:z-represent}

What precisely does the representation $\cz$ represent?
Given a prior $m(\nz \mid \cz, \ca)$ and latent-conditioned policy $\pi^z(\ca \mid \cz)$, we can use the current representation $\cz$ to predict good actions at both the current time step and (by unrolling the prior) at future time steps. Thus, \emph{the representation $\cz$ can be thought of as a compact representation of open-loop action sequences}.
Our experiments demonstrate that this compact representation of action sequences, once learned on one task, can be used to quickly learn a range of downstream tasks.

We can view RPC as \emph{learning a new action space for the MDP}. Precisely, define a new MDP where the actions are $\cz$ and the reward function is $\tilde{r}$ (Eq.~\ref{eq:reward-and-info}). What we previously called the encoder, $\phi(\cz \mid \cs)$, is now the policy for selecting the actions $\cz$ in this new MDP. The $-\log \phi(\cz \mid \cs)$ term in from $\tilde{r}$ is the entropy of this encoder, so the encoder is performing MaxEnt RL on this new MDP.
This new MDP encodes a strong prior for open-loop policies: simply sampling actions from the prior yields high reward.

\paragraph{Open-loop control.}
RPC learns a model $m(\cz \mid \mathbf{z_{t-1}}, \mathbf{a_{t-1}})$ that predicts the state representation at the next time step. Thus, we can unroll our policy in an open-loop manner, without observing transitions from the true system dynamics. Because the model and policy are trained to be self-consistent, we expect that the highly compressed policies learned by RPC will perform well in this open-loop setting, as compared to uncompressed policies (see experiments in Sec.~\ref{sec:experiments}).

\subsection{Theoretical Guarantees}

We conclude this section by formally relating model compression to open-loop performance and generalization error. We do not intend this section to be a comprehensive analysis of all benefits deriving from model compression (see, e.g., ~\citep{arora2018stronger, bassily2018learners}). All proofs are in Appendix~\ref{appendix:proofs}.

Intuitively, we know that a policy that uses zero bits of information will perform identically in the open-loop setting. The following result shows that our model compression objective corresponds to maximizing a lower bound on the expected return of the open-loop policy.
\begin{lemma} \label{lemma:open-loop}
Let encoder $\phi(\cz \mid \cs)$, policy $\pi^z(\ca \mid \cz)$, prior $m(\cz \mid \mathbf{z_{t-1}}, \mathbf{a_{t-1}})$, and reward function $r(\cs, \ca) > 0$ be given. Then our model compression objective (Eq.~\ref{eq:objective}) with reward function $(1 - \gamma) \log r(\cs, \ca)$ is a lower bound on the expected return of the open-loop~policy.
{\footnotesize \begin{equation*}
    \E_{\pi^\text{open}(\mathbf{\tau})}\left[\sum_{t=1}^\infty \gamma^t r(\cs, \ca) \right]
    \ge f \left( \E_{\pi^\text{reactive}(\mathbf{\tau})}\left[\sum_{t=1}^\infty \gamma^t \left((1 - \gamma) \log r(\cs, \ca) + \log m(\cz \mid \mathbf{z_{t-1}}, \mathbf{a_{t-1}}) - \log \phi(\cz \mid \cs) \right) \right] \right),
\end{equation*}}
where $f(x) = \frac{\gamma}{1 - \gamma}e^{\frac{x}{\gamma}}$ is a monotone increasing function of $x$.
\end{lemma}

Our next result shows that, not only does RPC optimize for open loop performance, but the difference between the performance of the open-loop policy and the reactive policy can be bounded by the policy's bitrate. This result could be useful for quantifying the regret incurred when using the learned representation $\cz$ as an action space of temporally-extended behaviors.

Let $\pi^\text{open}$ be the open-loop policy corresponding to the composition of the prior $m(\nz \mid \cz, \ca)$ and the high-level policy $\pi^z(\ca \mid \cz)$. Let $\pi^\text{reactive}$ be the reactive policy corresponding to the composition of the encoder $\phi(\cz \mid \cs)$ and the high-level policy $\pi^z(\ca \mid \cz)$. To simplify notation, we further define the sum of discounted rewards of a trajectory as $R(\mathbf{tau}) \triangleq \sum_t \gamma^t r(\cs, \ca)$ and let $R_\text{max} = \max_{\mathbf{\tau}} R(\mathbf{\tau})$ be the maximum return of any trajectory.
\begin{lemma} \label{lemma:open-loop-bound}
The expected return of the open-loop policy $\pi^\text{open}$ is at most $R_\text{max}\sqrt{\nicefrac{C}{2}}$ worse than the expected return of the reactive policy $\pi^\text{reactive}$:
\begin{equation*}
    \E_{\pi^\text{open}}\bigg[\sum_t \gamma^t r(\cs, \ca) \bigg] \ge \E_{\pi^\text{reactive}}\bigg[\sum_t \gamma^t r(\cs, \ca) \bigg] - R_\text{max}\sqrt{\nicefrac{C}{2}}.
\end{equation*}
\end{lemma}

One oft-cited benefit of compression is that the resulting models generalize better from the training set to the testing set. Our final result applies this same reasoning to RL. We assume that the policy has been trained on an empirical distribution of MDPs, and will be evaluated on a new MDP sampled from that sample distribution. For the following result, we assume that the given stochastic MDP is a mixture of deterministic MDPs, each described by a $b$-bit random string (see~\citet{ng2013pegasus}).
\begin{lemma}
Let stochastic MDP $\gM$ and reactive policy $\pi^\text{reactive}$ be given. Define $R^\pi(\mathbf{\gM})$ to be the expected reward of policy $\pi$ on MDP $\mathbf{\gM}$. Then the probability that the policy's expected return on an observed (deterministic) MDP $\hat{\gM}$ is much different than the policy's expected return on the stochastic MDP is bounded by the policy's bitrate:
\begin{equation*}
    P_{\hat{\mathbf{\gM}}}[|R^\pi(\hat{\mathbf{\gM}}) - R^\pi(\mathbf{\gM})| > \epsilon] \le \frac{C + 1}{2b \epsilon^2 - 1}.
\end{equation*}
\end{lemma}
The proof is a direct application of~\citet[Theorem 8]{bassily2018learners}.
This result may be of interest in the offline RL setting, where observed trajectories effectively constitute a deterministic MDP. The intuition is that policies that use fewer bits will be less likely to overfit to the offline dataset.

In summary, our theoretical results suggest that compressed policies learn representations that can be used for planning and that may generalize better.
We emphasize that the theoretical benefits of model compression have been well studied in the supervised learning literature. As our method is likewise performing model compression, we expect that it will inherit a wide range of additional guarantees, such as guarantees about sample complexity.

\section{Experiments}
\label{sec:experiments}

Our experiments have two aims. First, we will demonstrate that RPC achieves better compression than alternative approaches, obtaining a higher reward for the same number of bits. Second, we will study the empirical properties of compressed policies learned by our method, such as their robustness and ability to learn representations suitable for hierarchical RL. We do not intend this section to exhaustively demonstrate every possible benefit from compression; we acknowledge that there are many purported benefits of compression, such as exploration and sample efficiency, which we do not attempt to study here. We include additional experiments in Appendix~\ref{appendix:details}.

\begin{wraptable}[7]{r}{0.5\textwidth}
\vspace{-1em}
{\footnotesize
\begin{tabular}{c||p{1.5cm}|p{1cm}|p{0.9cm}}
 & feedforward architecture & predicted prior & augmented reward \\\midrule
RPC (ours) & \cmark & \cmark & \cmark \\
VIB~\citep{igl2019generalization} & \cmark & \xmark & \xmark \\
VIB+reward~\citep{lu2020dynamics} & \cmark & \xmark & \xmark \\
VIB+RNN & \xmark & \xmark & \cmark
\end{tabular}}
\end{wraptable}

\vspace{-0.3em}
\subsection{Evaluating Compression}
\vspace{-0.3em}

Our first experiment studies whether RPC  outperforms alternative approaches to compression, which we summarize in the inline table. We compare against a standard VIB~\citep{igl2019generalization} and an extension that adds the information cost to the reward, ``VIB+reward''~\citep{lu2020dynamics}. This baseline can be viewed as a special case of RPC where the blue arrows in Fig.~\ref{fig:pgm} are removed.
Finally, since the predictive prior in RPC is similar to a recurrent network, we compare against an extension of the VIB~\citep{igl2019generalization} that uses an LSTM (``RNN + VIB''). Unlike RPC, this baseline requires training on entire trajectories (rather than individual transitions) and requires backpropagating gradients through time. We evaluate all methods on four tasks from OpenAI-Gym~\citep{brockman2016openai} and two \emph{image-based} tasks from dm-control~\citep{tassa2018deepmind}. Because of computational constraints, we omit the RNN+VIB baseline on the image-based tasks.

\begin{figure}[t]
    \centering
    \vspace{-1.5em}
    \includegraphics[width=\linewidth]{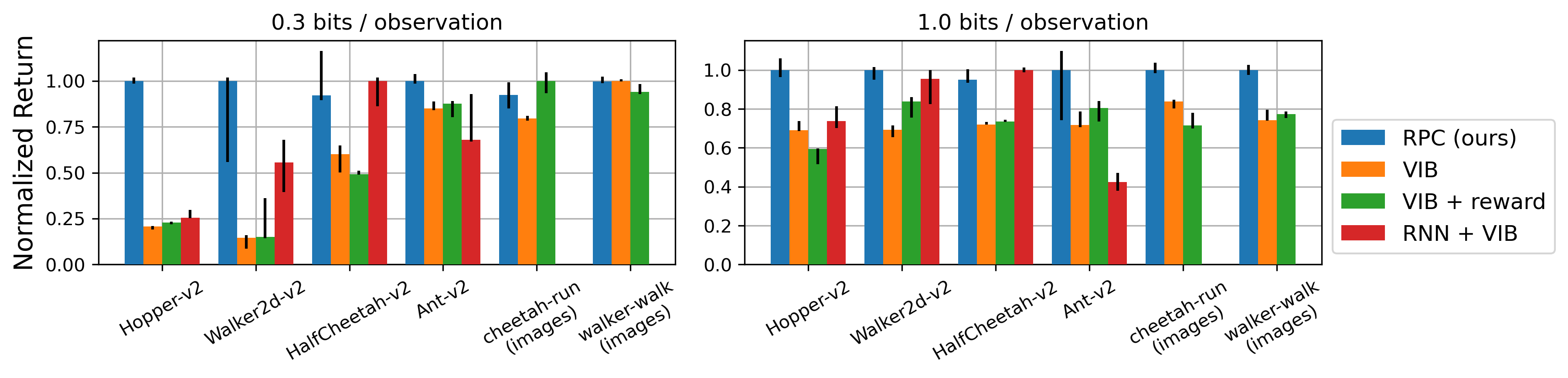}
    \vspace{-2em}
    \caption{{\footnotesize \textbf{Learning Compressed Policies.} We measure the return achieved by policies constrained to have a fixed bit rate. Especially at low bit rates, RPC achieves higher return than alternative methods.}}
    \label{fig:bits_vs_reward}
    \vspace{-0.5em}
\end{figure}

We plot results in Fig.~\ref{fig:bits_vs_reward}. To make the rewards comparable across tasks, we normalize the total return by the median return of the best method. On almost all tasks, RPC achieves higher returns than prior methods for the same bitrate. For example, when learning the Walker task using 0.3 bits per observation, RPC achieves a return of 3,562, the VIB+RNN baseline achieves a return of 1,978 (-44\%), and the other VIB baselines achieve a return of around 530 (-85\%). We include full results on a wider range of bitrates in Appendix Fig.~\ref{fig:bits_vs_reward_appendix}.
While, in theory, the VIB+RNN baseline could implement RPC internally, in practice it achieved lower returns, perhaps because of optimization challenges associated with training LSTMs~\citep{cho2014learning}. Even if the VIB+RNN baseline could implement the strategy learned by RPC, RPC is simpler (it does not require training on trajectories) and trains about 25\% times faster (it does not require backpropagation through time).

\begin{figure}[t]
    \centering
    \begin{minipage}{0.65\textwidth}
    \begin{subfigure}[b]{0.5\textwidth}
        \includegraphics[width=\linewidth]{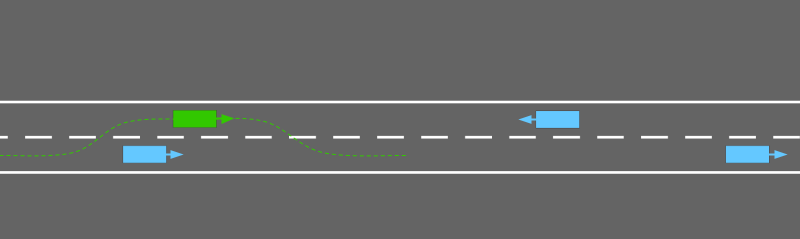}
        \includegraphics[width=\linewidth]{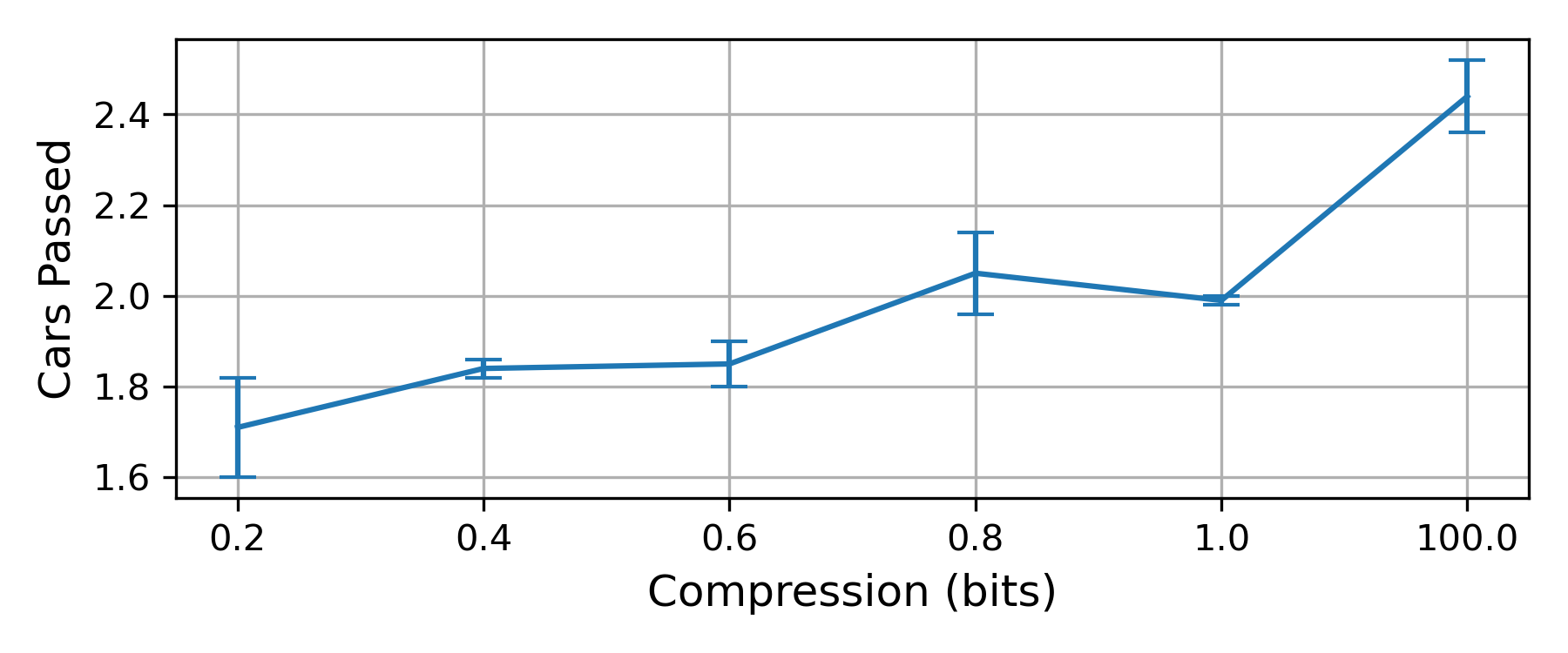}
        \caption{{\footnotesize Driving with Traffic}}
        \label{fig:two_way_qualitative}
    \end{subfigure}%
    ~
    \begin{subfigure}[b]{0.5\textwidth}
        \includegraphics[width=\linewidth, trim=0 20 0 0]{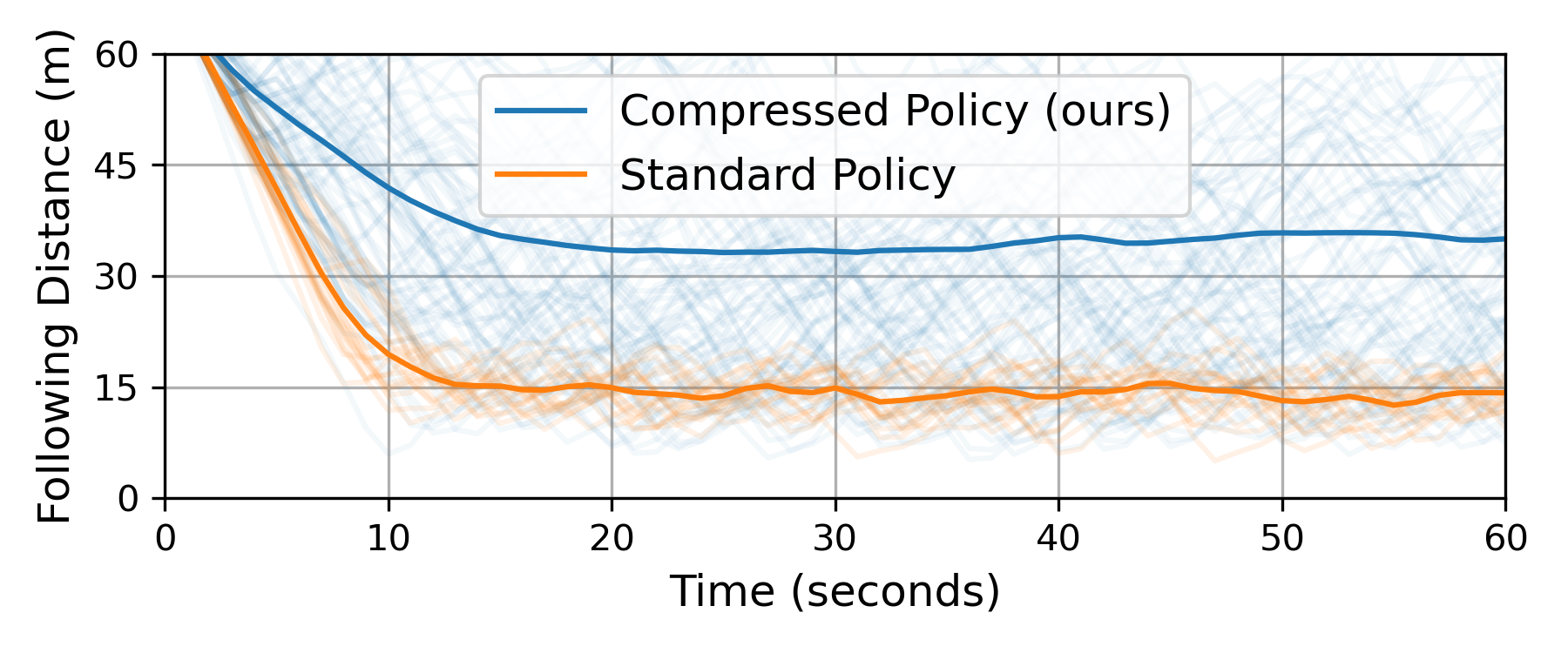}
        \includegraphics[width=\linewidth]{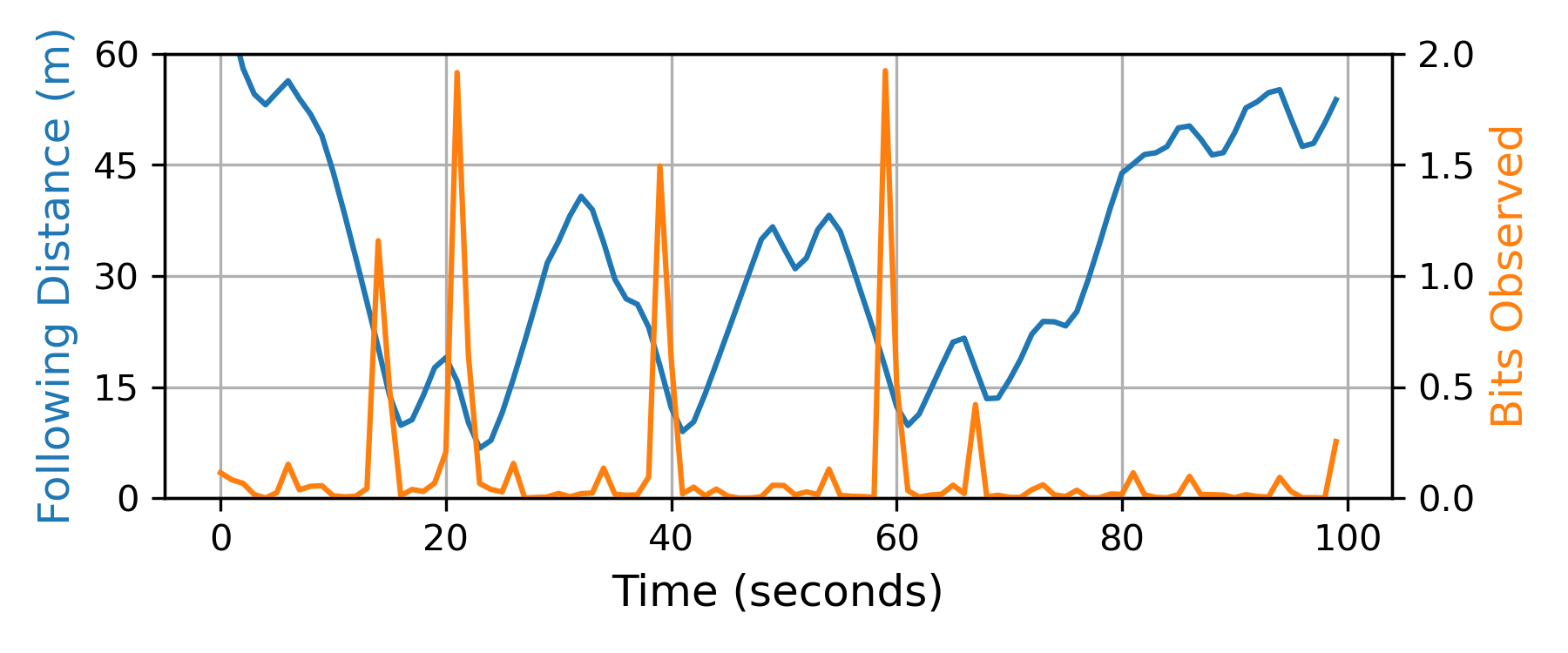} 
        \caption{{\footnotesize Active Cruise Control}}
        \label{fig:acc_qualitative}
    \end{subfigure}
    \caption{{\footnotesize \textbf{Behavior of compressed policies}: On two driving tasks, we observe that highly compressed policies \figleft \, avoid passing other cars and \figright \, leave a larger following distance between cars. The passing and tailgating that our method forgoes would require more bits of information about the precise locations of the other cars.}
    \label{fig:behavior}
    }
    \end{minipage}
    \hfill
    \begin{minipage}{0.3\textwidth}
        \includegraphics[width=\linewidth]{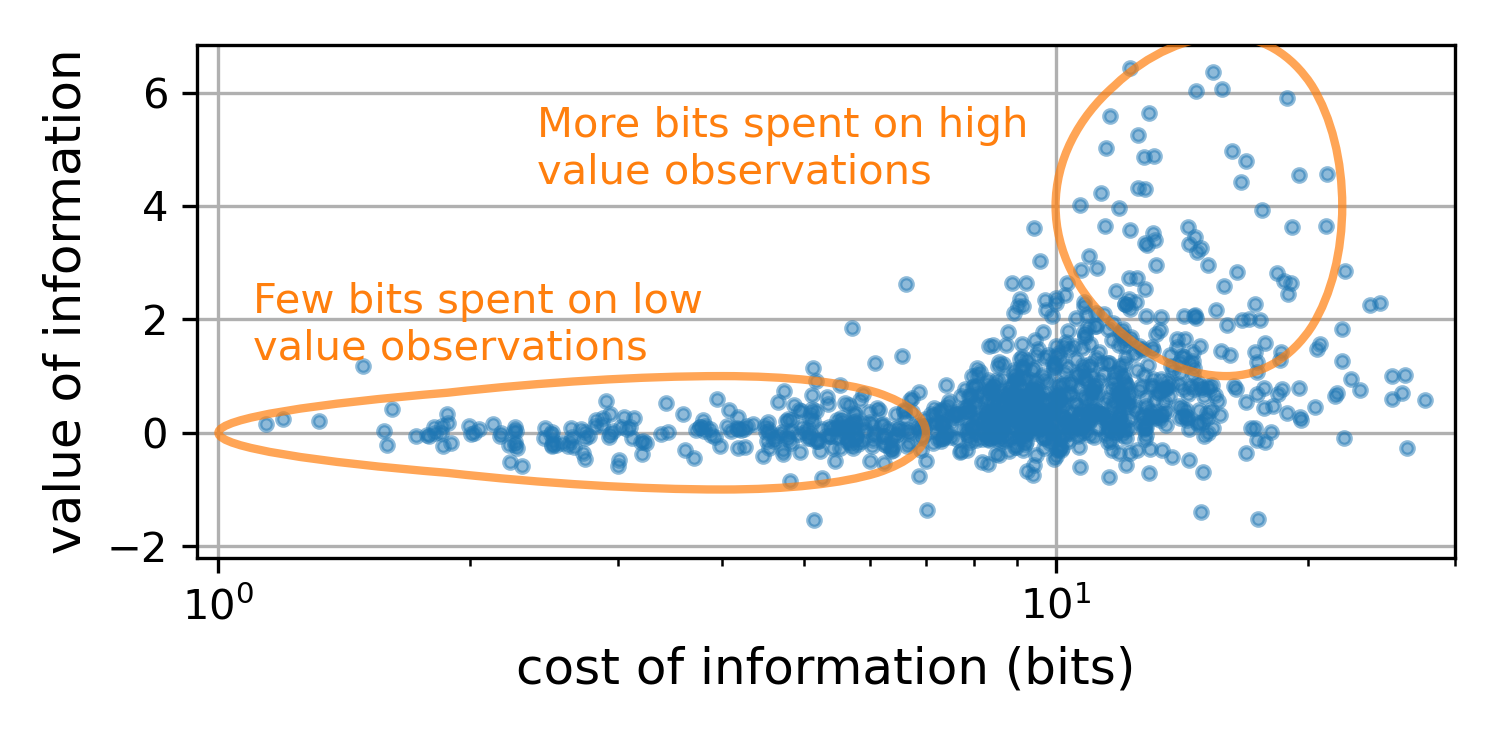}
        \includegraphics[trim=0 0 0 30, width=\linewidth]{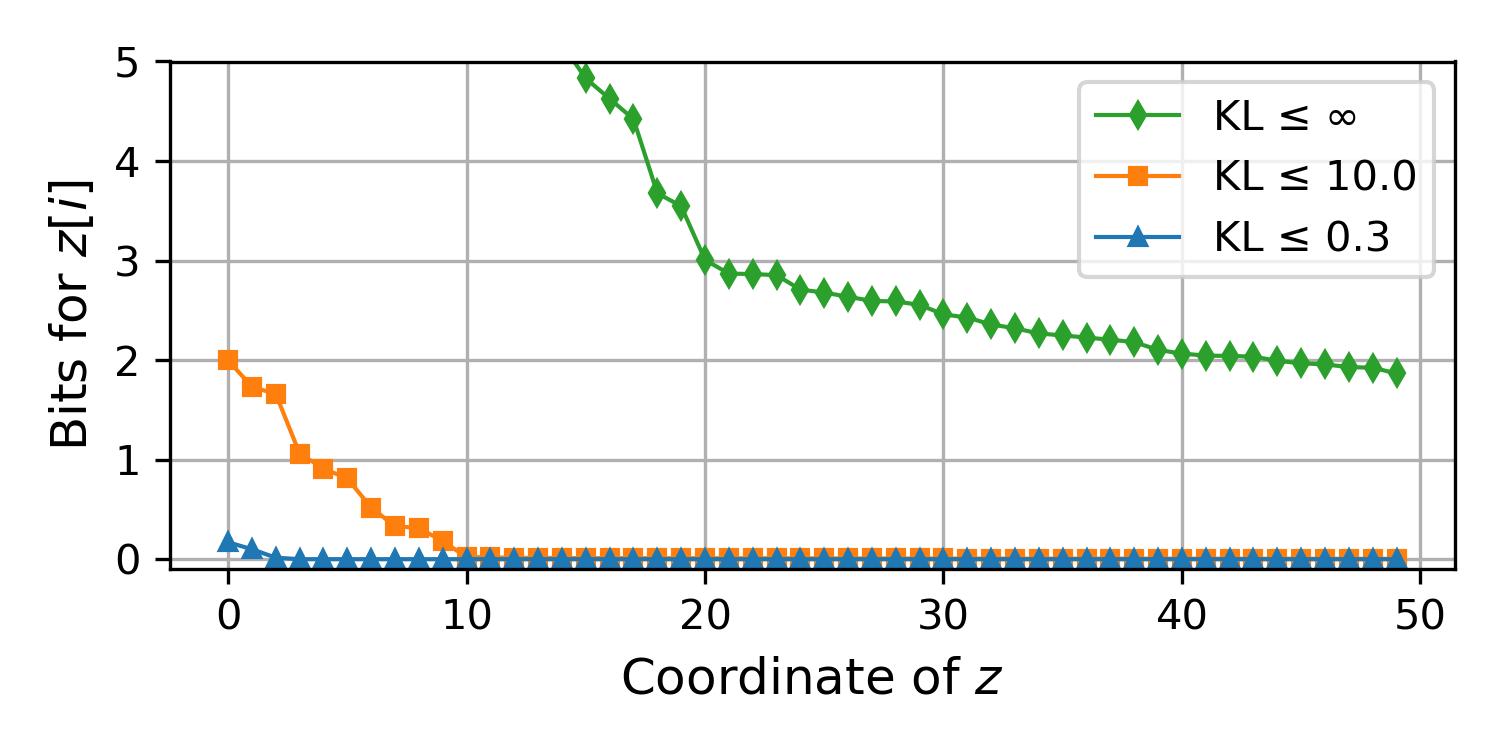}%
    \caption{{\footnotesize \textbf{Representations}: \figtop \;
    Compressed policies observe more bits from observations that have a large value of information.
    \figbottom \; Compressed policies learn sparse representations.
    }} \label{fig:representations}
    \end{minipage} 
    \vspace{-1em}
\end{figure}

\vspace{-0.3em}
\subsection{Visualizing Compressed Policies}
\vspace{-0.3em}

\paragraph{Behavior of compressed policies.}
To visualize how compression changes behavior, we applied RPC to two simulated driving tasks shown in Fig.~\ref{fig:two_way_qualitative} (top left), which are based on prior work~\citep{highwayenv}.
In the first task, the agent can pass cars by driving into the lane for oncoming traffic; the second task uses the same simulated environment but restricts the agent to remain in its own lane.
The rewards for these tasks corresponds to driving to the right as quickly as possible, without colliding with any other vehicles.
In the first task, we observe that compressed policies passed fewer cars than uncompressed policies. Intuitively, a passing maneuver requires many bits of information about the relative positions of other cars, so we expect that compressed policies to engage in fewer passing maneuvers. Fig.~\ref{fig:two_way_qualitative} (bottom) shows that the bitrate of RPC is directly correlated with the number of cars passed.
In the second task, we observe that compressed policies leave a larger following distance from the leading car (Fig.~\ref{fig:acc_qualitative} (top)). Fig.~\ref{fig:acc_qualitative} (bottom) shows that the number of bits used per observation increases when the car is within 15m of another car. Thus, by maintaining a following distance of more than 30m, the compressed policy can avoid these situations where it would have to use more bits.
See the project website for videos.\footnote{Project site with videos and code: \url{https://ben-eysenbach.github.io/rpc}}

\paragraph{Representations of compressed policies.}
A policy learned by compression must trade off between maximizing reward and paying to receive bits of information. Using the HalfCheetah task, we plot the value of information versus cost of information for many sampled states. See Appendices~\ref{sec:voi} and~\ref{sec:voi-experiment} for details.
As shown in Fig.~\ref{fig:representations} (top), the RPC uses more bits from observations that are more valuable for maximizing future returns.
Then, to visualize the learned representation $\cz$, we sample the representation from randomly sampled states, and plot the numbers of bits used for each coordinate. Fig.~\ref{fig:representations} (bottom) shows that RPC learns sparse representations. Whereas the uncompressed policy uses all coordinates, a policy compressed with bitrate 10 uses only $\nicefrac{10}{50}$ coordinates and a policy compressed with bitrate 0.3 uses only $\nicefrac{2}{50}$ coordinates.

\vspace{-0.3em}
\subsection{Hierarchical RL using the Learned Representation}
\vspace{-0.3em}

\begin{figure}[t]
    \vspace{-1.5em}
    \centering
    \begin{subfigure}[c]{0.3\textwidth}
        \includegraphics[width=0.49\linewidth]{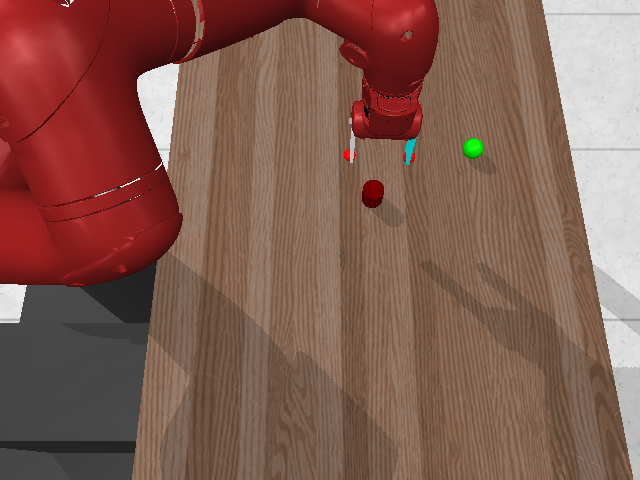}
        \includegraphics[width=0.49\linewidth]{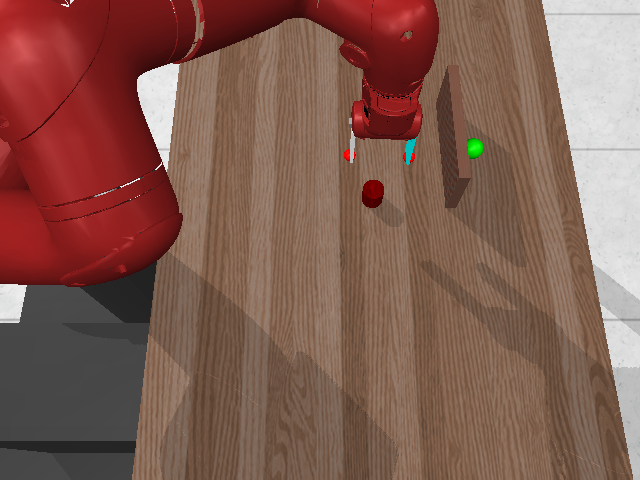}
        \includegraphics[width=0.49\linewidth]{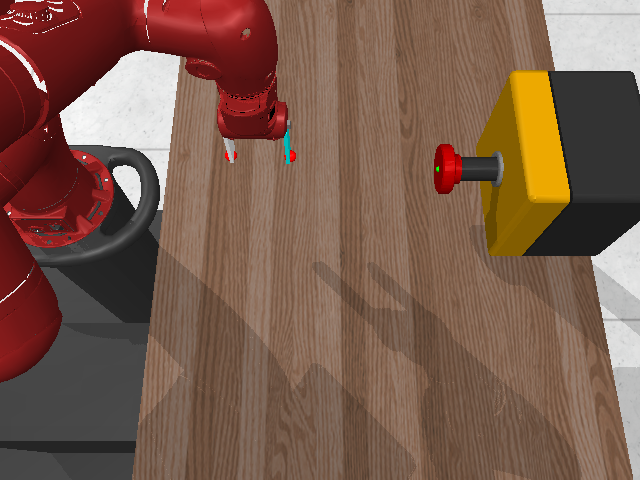}
        \includegraphics[width=0.49\linewidth]{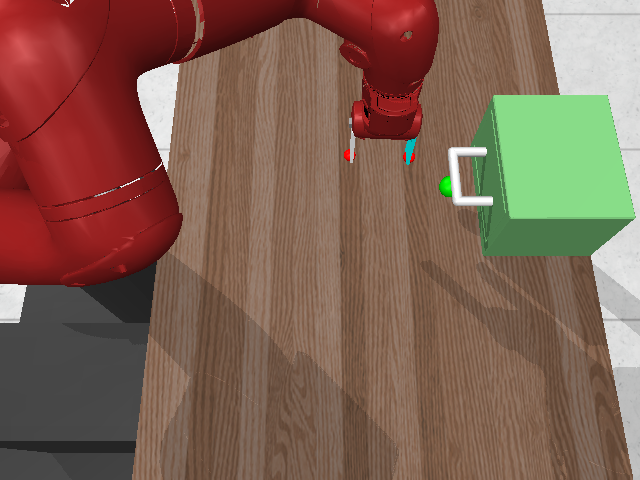}
        
    \end{subfigure}%
    ~
    \begin{subfigure}[c]{0.7\textwidth}
        \includegraphics[width=\linewidth]{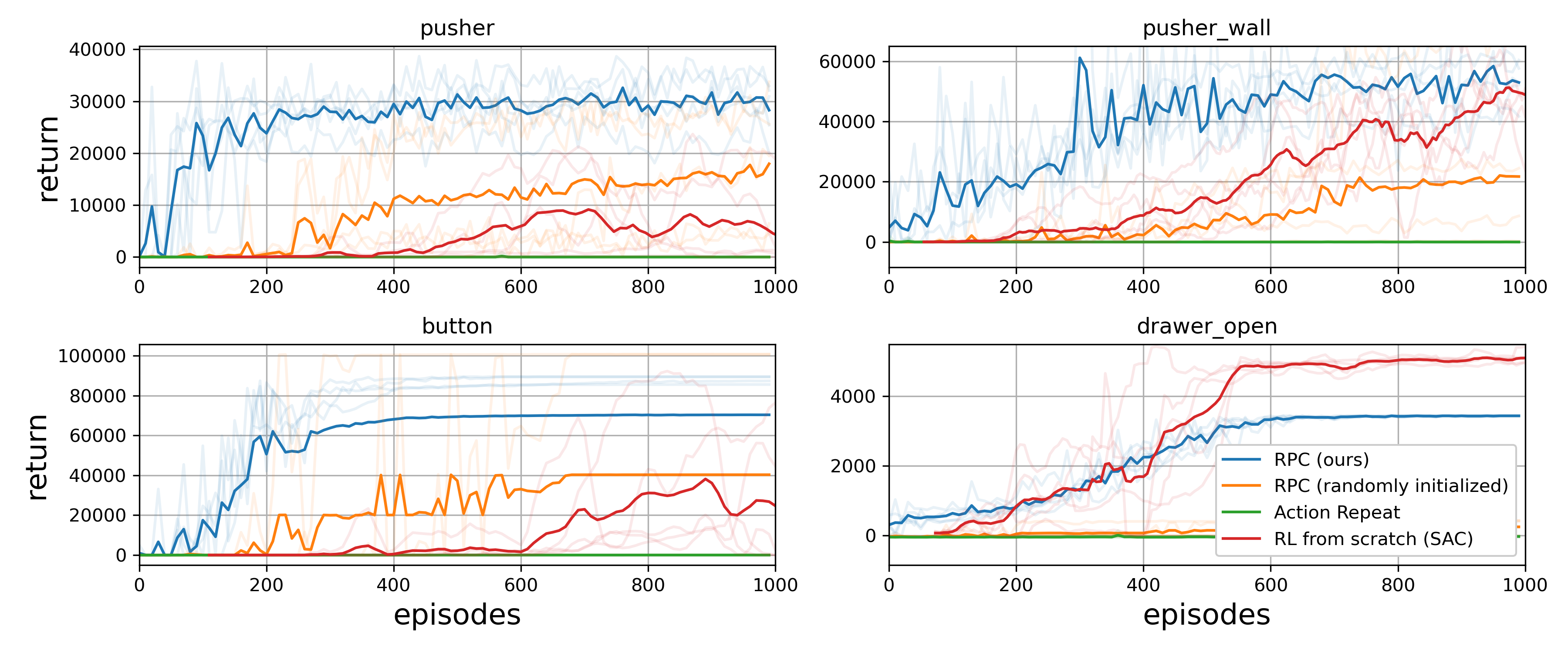}
    \end{subfigure}
    \vspace{-1em}
    \caption{{\footnotesize \textbf{Hierarchical RL}: We apply RPC to the pushing task shown in the top-left. We then use the learned representation of action sequences as the action space for solving three new tasks. On all tasks, the representations learned by RPC accelerate learning. All episodes have a fixed length.}}
    \label{fig:hrl}
    \vspace{-1.5em}
\end{figure}

RPC learns a representation of temporally-extended action sequences (see Sec.~\ref{sec:z-represent}).
We hypothesize that these representations can accelerate the learning of new tasks in a hierarchical setting.
Our goal is not to propose a complete hierarchical RL system, but rather evaluate whether these action representations are suitable for high-level control.
During training, we apply RPC to a goal-conditioned object pushing task, shown in Fig.~\ref{fig:hrl} (top-left). The initial position of the object and the goal position are randomized, so we expect that different representations $\cz$ will correspond to high-level behaviors of moving the end effector and object to different positions. At test-time, the agent is presented with a new task. The agent will attempt to solve the task by commanding one or two behaviors $\cz$. See Appendix~\ref{appendix:details} for details and pseudocode.

We compare the action representations learned by RPC to three baselines. To test whether RPC has learned a prior over \emph{useful} action sequences, we use a variant of RPC (``RPC (randomly initialized)'' ) where the policy and model are randomly initialized. ``Action Repeat'' constantly outputs the same action. Finally, ``RL from scratch'' applies a state-of-the-art off-policy RL algorithm (SAC~\citep{haarnoja2018soft}) to the task. 
We apply all methods to four tasks and present results in Fig.~\ref{fig:hrl}. First, as a sanity check, we apply all methods to the training task, finding that RPC quickly finds a single $\cz$ that solves the task. On the remaining three tasks, we likewise observe that the representations learned by RPC allow for much faster learning than the baselines. The final task, ``pusher wall'' requires chaining together multiple representations $\cz$ to solve. While the ``RL from scratch'' baseline eventually matches and then surpasses the performance of RPC, RPC accelerates learning in the low-data regime.

\vspace{-0.3em}
\subsection{Robustness}
\label{sec:experiments-robust}
\vspace{-0.3em}

The connection between compression and robustness has been well established in the literature (e.g.,~\citep{ye2019adversarial, gui2019model}). Our next set of experiments test the robustness of compressed policies to different types of disturbances: missing observations (i.e., open-loop control), adversarial perturbations to the observations, and perturbations to the dynamics (i.e., robust RL).
We emphasize that since these experiments focus on robustness, the policy is trained and tested in different environments.

\begin{wrapfigure}[9]{r}{0.55\textwidth}
    \centering
    \vspace{-1em}
    
    \includegraphics[width=0.5\linewidth]{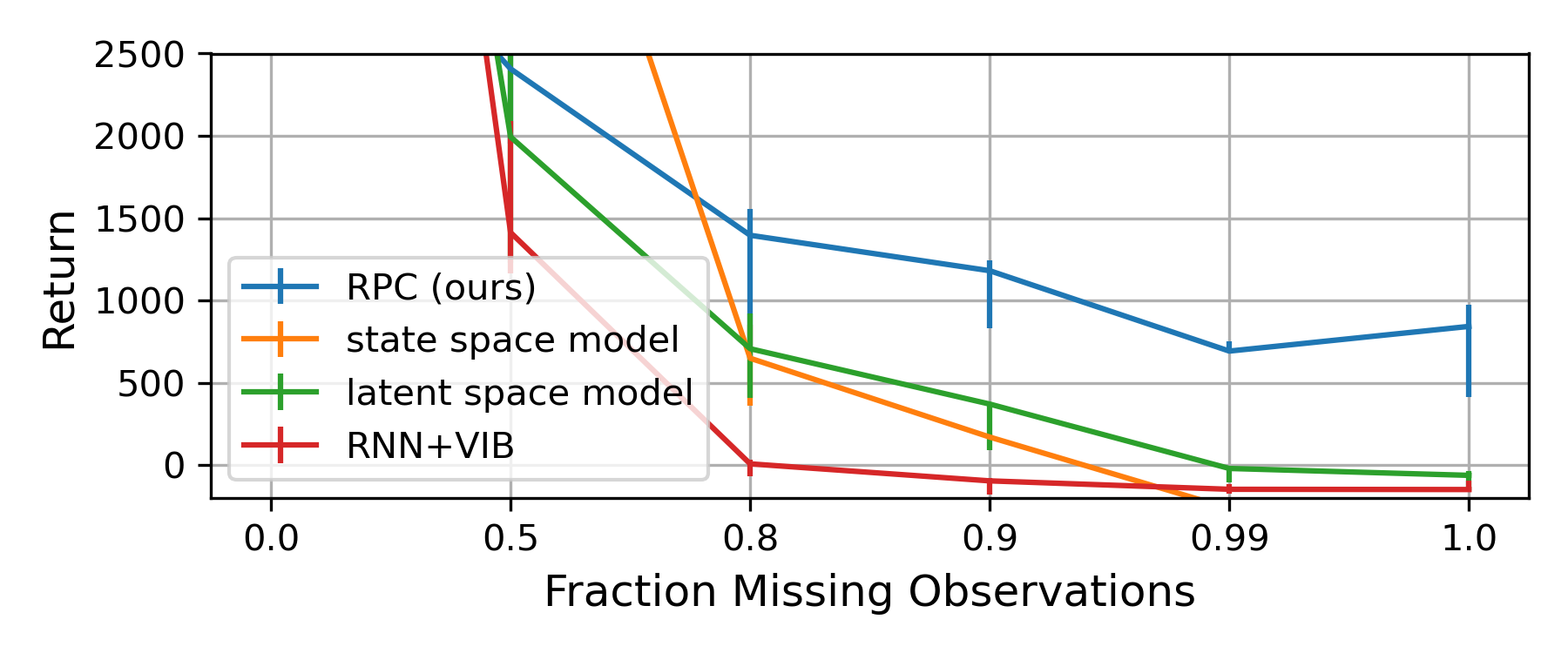}%
    ~
    \includegraphics[width=0.5\linewidth]{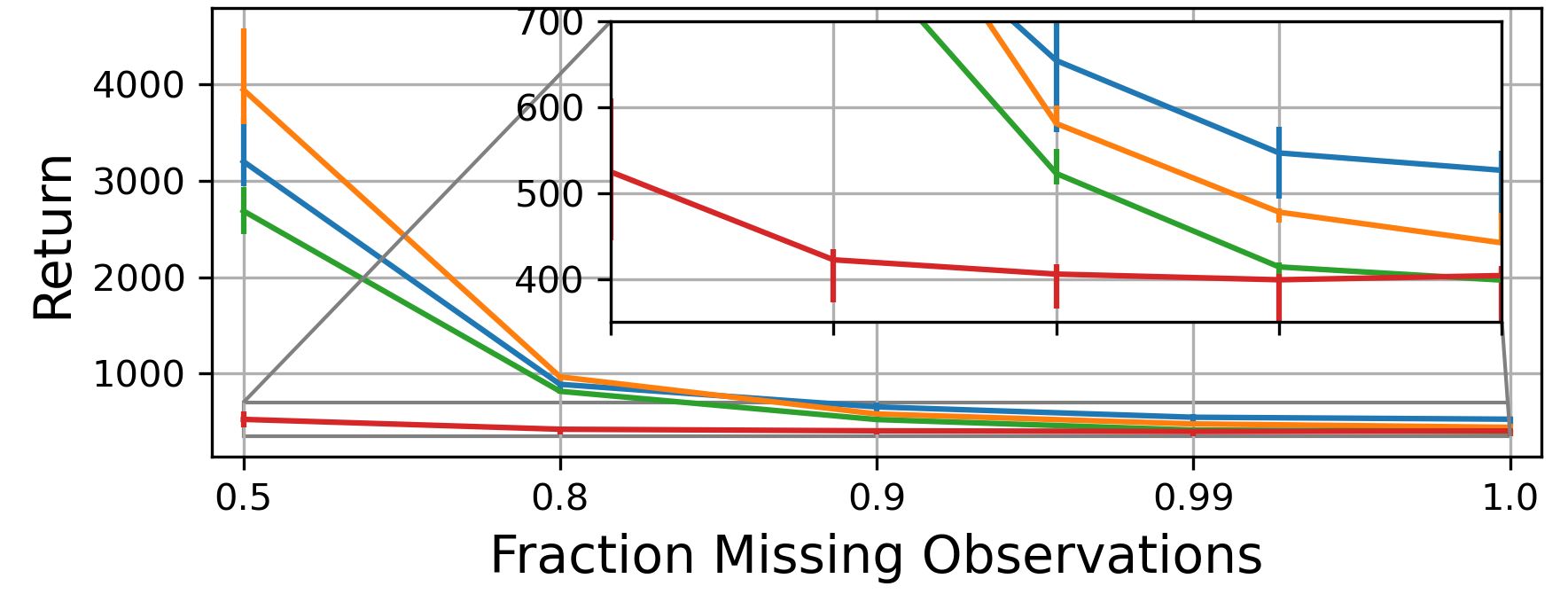}
    
    \caption{{\footnotesize \textbf{Robustness to missing observations}: RPC is more robust to missing observations than prior methods, including those that learn dynamics models. We show \texttt{HalfCheetah-v2} on left and \texttt{Walker2d-v2} on right.}}
    \label{fig:sensor_dropout}
\end{wrapfigure}
\paragraph{Robustness to missing observations and open-loop control.}
Since compressed policies rely on fewer bits of input from the observations, we expect that they not only will be less sensitive to missing observations, but will actively modify their behavior to adopt strategies that require fewer bits from the observation. In this experiment, we drop each observation independently with probability $p \in [0, 1]$, where $p = 1$ corresponds to using a fully open-loop policy. For RPC, we handle missing observations by predicting the representation from the previous time step. Our two main baselines take a policy used by standard RL and learn either a latent-space model or a state-space model. When the observation is missing, these baselines make predictions using the learned model. We also compare against RNN+VIB, the strongest baseline from Fig.~\ref{fig:bits_vs_reward}. When observations are missing, the LSTM's input for that time step is sampled from the prior. Fig.~\ref{fig:sensor_dropout} shows that all methods perform similarly when no observations are dropped, but RPC achieves a higher reward than baselines when a larger fraction of observations are dropped. This experiment shows that more effective compression, as done by RPC, yields more robust policies. This robustness may prove useful in real-world environments where sensor measurements are missing or corrupted.

\paragraph{Adversarial Robustness.}

Compressed policies extract fewer bits from each observation, so we expect that compressed policies will be more robust to adversarial perturbations to the observation.
While prior work has proposed purpose-designed methods for achieving robustness~\citep{morimoto2005robust, tessler2019action}, here we investigate whether compression is simple yet effective means for achieving a modicum of robustness. We do not claim that compression is the best method for learning robust policies.

\begin{wrapfigure}[8]{R}{0.5\textwidth}
    \vspace{-1.5em}
    \centering
        \includegraphics[width=0.5\linewidth]{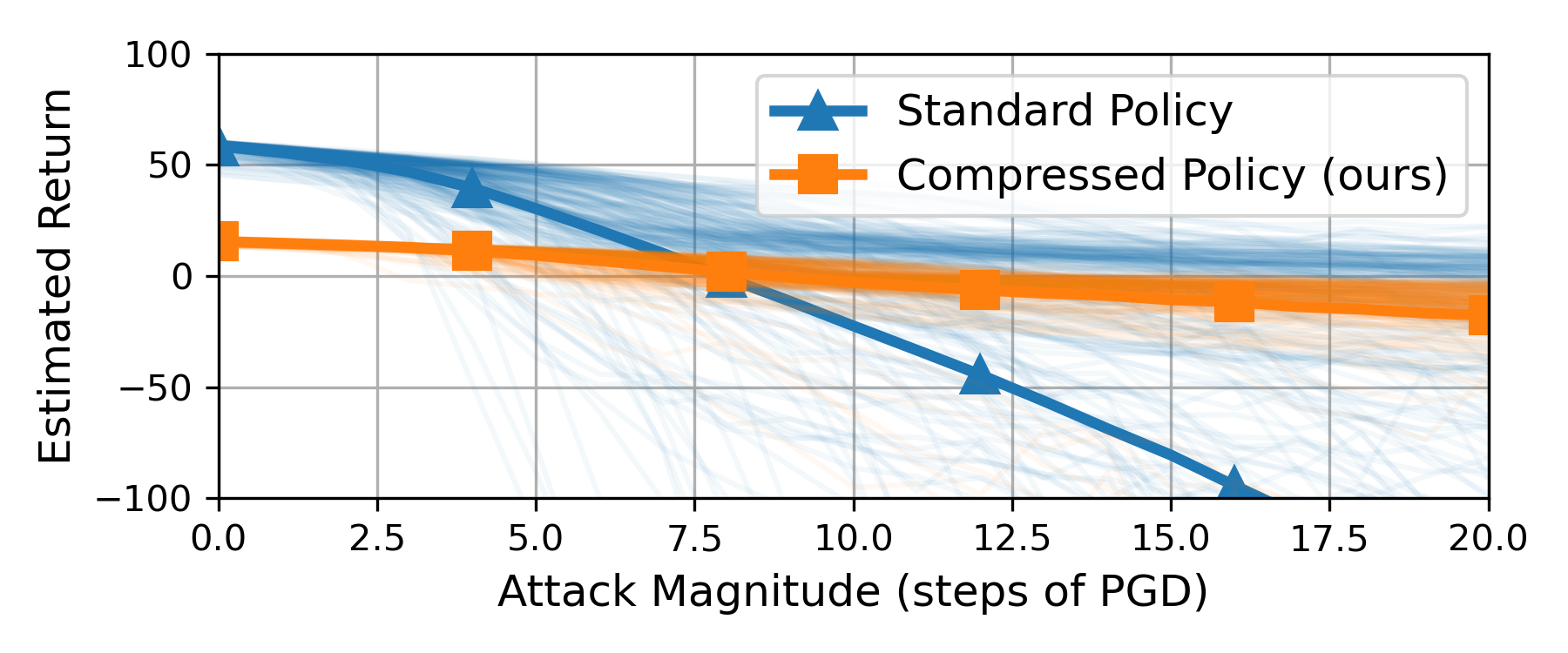}
        \includegraphics[width=0.45\linewidth]{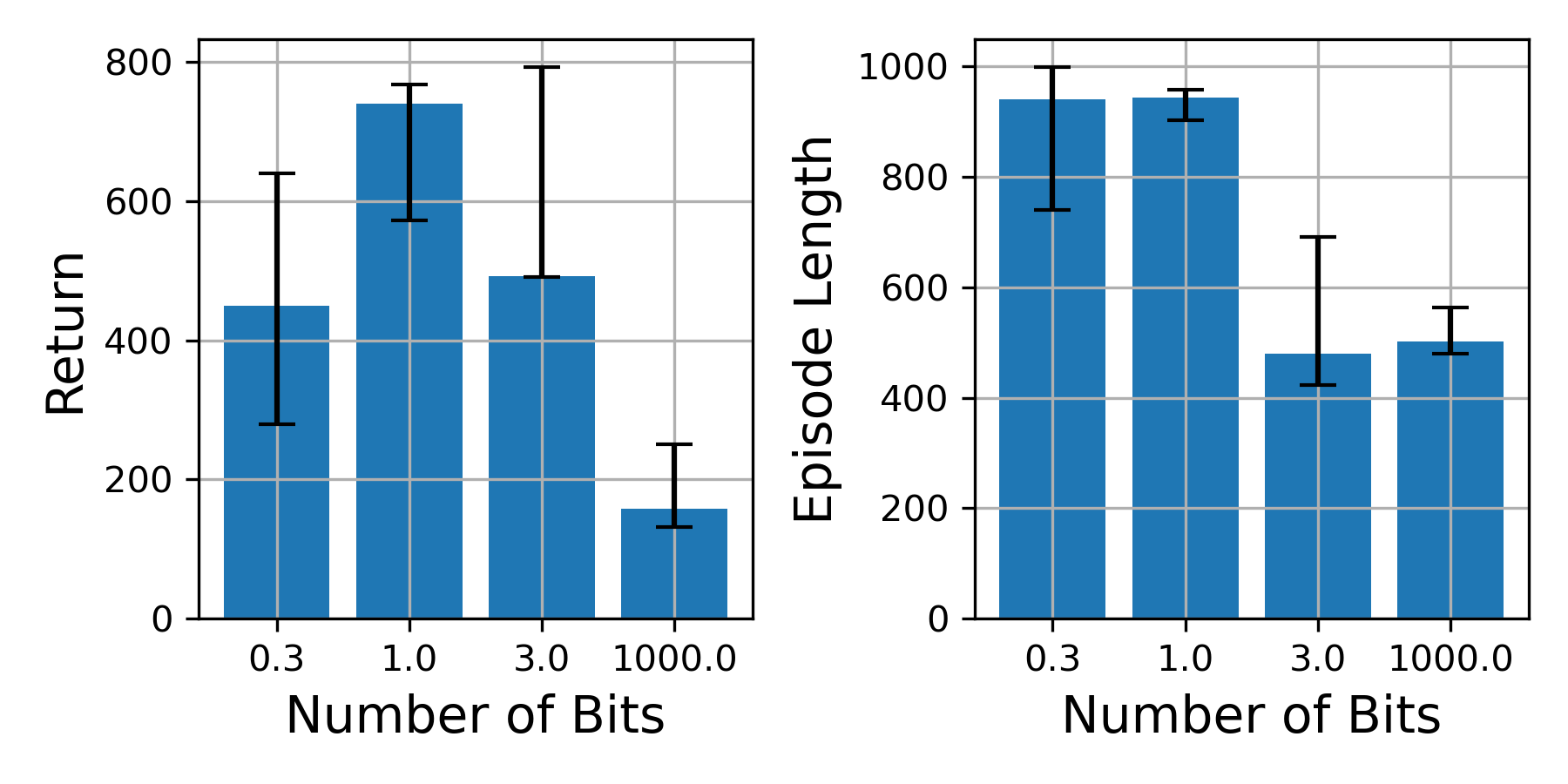}
    \vspace{-0.5em}
    \caption{{\footnotesize \textbf{Adversarial robustness}: Compressed policies are more robust against adversarial attacks to the \figleft \, dynamics and \figright \, observations.}}
    \label{fig:pgd}
\end{wrapfigure}

We first study adversarial perturbations to the \textbf{dynamics} on the \texttt{Ant-v2} environment.
Given a policy $\pi(\ca \mid \cs)$ and the current state $\cs$, the adversary aims to apply a small perturbation to that state to make the policy perform as poorly as possible. We implement the adversary using projected gradient descent~\citep{madry2017towards}; see Appendix~\ref{appendix:details} for details. Fig.~\ref{fig:pgd} (left) shows the expected return %
as we increase the magnitude of the attack. The compressed policy is more resilient to larger attacks than the uncompressed policy. 
Our next experiment looks at perturbations to the \textbf{observation}. Unlike the previous experiment, we let the adversary perturb \emph{every} step in an episode; see Appendix~\ref{appendix:details} for full details.
Fig.~\ref{fig:pgd} (right) shows that policies that use fewer bits achieve higher returns in this adversarial setting. When the \texttt{Ant-v2} agent uses too many bits (3 or more), the adversarial perturbations to the dynamics flip the agent over, whereas agents that use fewer bits remain upright.

\paragraph{Robust RL.}

\begin{wrapfigure}[9]{R}{0.5\textwidth}
    \centering
    \vspace{-1.5em}
    \includegraphics[width=\linewidth]{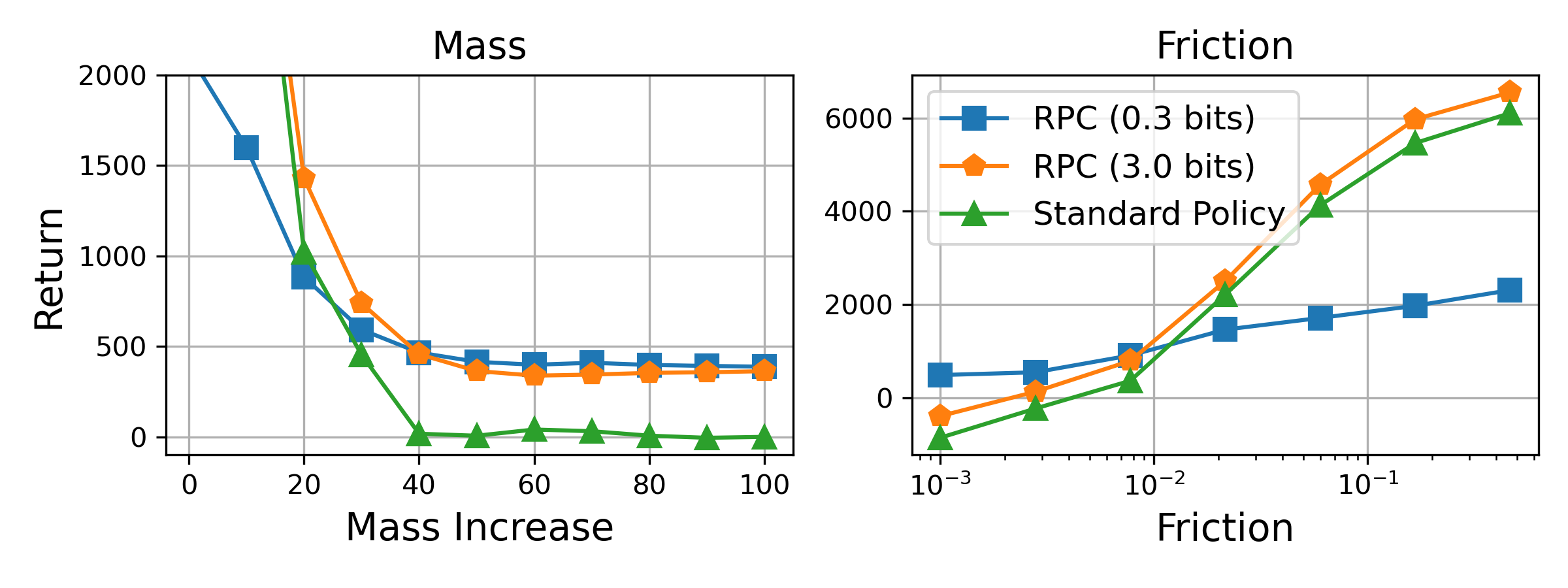}
    \vspace{-1.8em}
    \caption{{\footnotesize \textbf{Robust RL}: Compressed policies are more robust to increases in mass and decreases in friction.}}
    \label{fig:robust-rl}
\end{wrapfigure}
Our final set of experiments look at higher-level perturbations to the dynamics, as are typically studied in the robust RL community~\citep{morimoto2005robust, tessler2019action}. Using the same \texttt{Ant-v2} environment as before, we (1) increase the mass of each body element by a fixed multiplier, or (2) decrease the friction of each body geometry by a fixed multiplier. These experiments test whether the learned policies are robust to more massive robots or more ``slippery'' settings. Fig.~\ref{fig:robust-rl} shows that compressed policies generalize to larger masses and smaller frictions more effectively than an uncompressed policy. Comparing the policies learned by RPC with a bit rate of 0.3 bits versus 3.0 bits, we observe that the bit rate effectively balances performance versus robustness.
See Appendix Fig.~\ref{fig:robust-rl-appendix} for a larger version of this plot with error bars.

\section{Conclusion}
\label{sec:conclusion}

In this paper, we presented a method for learning robust and predictable policies. Our objective differs from prior work by compressing \emph{sequences} of observations, resulting in a method that jointly trains a policy and a model to be self-consistent.
Not only does our approach achieve better compression than prior methods, it also learns policies that are more robust (Fig.~\ref{fig:pgd}). We also demonstrate that our method learns representations that are suitable for use in hierarchical RL.

\paragraph{Limitations.}
The main limitation of this work is that policies that use few bits will often receive lower reward on the training tasks. Also, for the purpose of exploration, the most informative states may be those that are hardest to compress. While the first limitation is likely irreconcilable, the second might be lifted by \emph{maximizing} information collected during exploration but \emph{minimizing} information for policy optimization.

\vspace{3em}
{\footnotesize
\paragraph{Acknowledgments.}
We thank Dibya Ghosh and the members of the Salakhutdinov lab for feedback throughout the project. We thank Oscar Ramirez for help setting up the image-based experiments and thank Rico Jonschkowski for introductions to potential collaborators. We thank Ryan Julian and Vincent Vanhoucke for feedback on the paper draft, and thank Abhishek Gupta for discussions during the early stages of this project.

This material is supported by the Fannie and John Hertz Foundation and the NSF GRFP (DGE1745016). %
}

\clearpage
\appendix

\section{More Analysis}

\subsection{Objective for the Encoder, Model, and Policy}
\label{appendix:one-step-obj}
This section describes how the objective for the encoder, model, and policy (Eq.~\ref{eq:one-step-obj}) is derived from our overall objective (Eq.~\ref{eq:main-obj}). Our aim is to maximizing the sum of (information-augmented) rewards (Eq. 3), starting at state $\cs$:
\begin{align*}
\mathcal{L}(\theta; \cs) 
&= E\left[ \sum_{t'=t}^\infty \gamma^{t' - t} \tilde{r}(\mathbf{s_{t'}}, \mathbf{a_{t'}}) \mid \cs \right] \\
& = E\left[ \sum_{t'=t}^\infty \gamma^{t' - t} (r(\mathbf{s_{t'}}, \mathbf{a_{t'}}) + \lambda ( \log m_\theta(\mathbf{z_{t'+1}} \mid \mathbf{z_{t'}}, \mathbf{a_{t'}}) - \log \phi_\theta(\mathbf{z_{t'}} \mid \mathbf{s_{t'}}))) \mid \cs \right] \\
&= E\left[ r(\cs, \ca) + \lambda ( \log m_\theta(\nz \mid \cz, \ca) - \log \phi_\theta(\cz \mid \cs)) \right. \\
& \qquad \qquad \left. + \sum_{t'=t+1}^\infty \gamma^{t' - t} (r(\mathbf{s_{t'}}, \mathbf{a_{t'}}) + \lambda ( \log m_\theta(\mathbf{z_{t'+1}} \mid \mathbf{z_{t'}}, \mathbf{a_{t'}}) - \log \phi_\theta(\mathbf{z_{t'}} \mid \mathbf{s_{t'}}))) \mid \cs \right] \\
&= E\left[ r(\cs, \ca) + \lambda ( \log m_\theta(\nz \mid \cz, \ca) - \log \phi_\theta(\cz \mid \cs)) + \gamma Q_\psi(\ns, \na) \right].
\end{align*}
Note that the reward at the current time step, $r(\cs, \ca)$, is not influenced by the parameters $\theta$, so we can drop this term:
\begin{equation*}
\mathcal{L}(\theta; \cs) \stackrel{\text{const.}}{=} E\left[\lambda ( \log m_\theta(\nz \mid \cz, \ca) - \log \phi_\theta(\cz \mid \cs)) + \gamma Q_\psi(\ns, \na) \right].
\end{equation*}
The remaining difference between this objective and Eq. 5 is that the Q value term is scaled by $\gamma$. However, this difference has no effect on the optimization problem because the parameter $\lambda$ is automatically tuned. If we scale the second term, $Q$, by some value (say, $\gamma$), then tuning $\lambda$ to satisfy the bitrate constraint will result in a different value for $\lambda$ (one which is $\gamma$ times smaller).
For optimizing this objective, we sample states $\cs$ from the replay buffer.

\subsection{Prior on the Initial Representation}
For simplicity, we have omitted the prior $m(\mathbf{z_1})$ on the representation of the first observation. This prior cannot be predicted from prior observations. Instead, we fix $m(\mathbf{z_1})$ to be a zero-mean, unit-variance Gaussian. The information-augmented reward at the first time step is therefore
\begin{equation*}
    \tilde{r}_\lambda(\mathbf{s_{0}}, \mathbf{a_{0}}, \mathbf{s_1}) \triangleq \cancel{r(\mathbf{s_0}, \mathbf{a_0})} + \lambda \left( \log m_\theta(\mathbf{z_1}) - \log \phi_\theta(\mathbf{z_1} \mid \mathbf{s_1}) \right).
\end{equation*}

\subsection{MaxEnt RL and the Optimal Encoder}
\label{appendix:optimizing-encoder-maxent}

Maximum entropy (MaxEnt) RL is a special case of our compression objective. If the high-level policy is the identity function ($\pi^z(\ca \mid \cz) = \delta(\ca = \cz)$) and the prior is the uniform distribution over $\gZ$ (i.e., $m(\nz \mid \cz, \ca) = \text{Unif}(\nz)$), then we recover standard MaxEnt RL. Another way of describing this connection is that MaxEnt RL is equivalent to imposing an information bottleneck on the final (i.e., action) output of the policy. This connection suggests that some of the empirically-observed benefits of MaxEnt RL might be derived from the fact that it implicitly is performing model compression. For example, the objective from MaxEnt RL can be interpreted as optimizing for behavior under which random actions (sampled from the prior) do not significantly decrease the expected reward. Nonetheless, directly optimizing for the model compression objective yields a prior that depends on time. As shown in our experiments, such a prior yields a policy that not only is more compressed, but is also more robust to sensor failure (i.e., the open-loop setting).

In MaxEnt RL, the optimal policy can be expressed in terms of the \emph{soft} value function: $\pi(\ca \mid \cs) \propto e^{\tilde{Q}(\cs, \ca)}$, where $\tilde{Q}(\cs, \ca)$ is the expected, entropy-regularized return. We can express the optimal encoder learned by RPC in similar terms.
First, we reparametrize the Q function in terms of $\cz$ instead of $\ca$: $Q^\pi(\cs,  \cz) = \E_{\pi^z(\ca \mid \cz)}[Q(\cs, \ca)]$.
The optimization problem for the encoder is
\begin{align*}
    \max_{\phi(\cz \mid \cs)} & \E_{\phi(\cz \mid \cs)}\left[\sum_t \gamma^t \tilde{r}(\cs, \ca, \ns) \right] \\
    &  = \E_{\phi(\cz \mid \cs)}\left[r(\cs, \ca) + \log m(\cz \mid z_{t-1}, a_{t-1}) - \log \phi(\cz \mid \cs) + \gamma Q(\ns, \na) \right].
\end{align*}
Using calculus of variations, we determine that the optimal encoder is given by
\begin{equation*}
    \phi(\cz \mid \cs) =  \frac{m(\cz \mid \mathbf{z_{t-1}}, \mathbf{a_{t-1}})e^{Q^\pi(\cs, \cz)}}{\int m(\cz' \mid \mathbf{z_{t-1}}, \mathbf{a_{t-1}})e^{Q^\pi(\cs, \cz')} d\cz'}.
\end{equation*}
Thus, the optimal encoder is trained to \emph{tilt} the predictions from the prior by the Q function.

\subsection{Value of information.}
\label{sec:voi}

An optimal agent must balance these information costs against the \emph{value of information} gained from these observations. Precisely, the value of information is how much more reward an optimal agent could receive if it observes the representation $\cz$ instead of predicting $\cz$ from the previous representation and action. We expect that the optimal policy will only look at representations where the value of information is greater than the cost of information. We confirm this prediction experimentally in Fig.~\ref{fig:representations}.
In practice, the policy learned by RPC looks at every observation, but may only look at a few bits from that observation.

\section{Experimental Details}
\label{appendix:details}

\begin{algorithm}[t]
\caption{\textbf{Robust Predictable Control}. The updates below are written using a learning rate of $\eta$. In practice we perform gradient steps using the Adam~\citep{kingma2015adam} optimizer. \label{alg:rpc}}
\begin{algorithmic}[1]
\State Initialize prior $m_\theta(\nz \mid \cz, \ca)$, encoder $\phi_\theta(\nz \mid \cs)$, Q function $Q_\psi(\cs, \ca)$, and high-level policy $\pi_\theta^z(\ca \mid \cz)$.
\State Initialize replay buffer $\gD \gets \emptyset$
\State Initialize dual variable $\log \lambda \gets \log(1e-6)$
\While{not converged}
    \State Sample a batch of transitions: $\{(\cs, \ca, \mathbf{r_t}, \ns) \sim \gD\}$
    \State Compute information cost: $\mathbf{c_t} \gets \E[\log \phi_\theta(\cz \mid \cs) - \log m_\theta(\nz \mid \cz, \ca)]$  \Comment{Eq.~\ref{eq:reward-and-info}}
    \State Compute information-regularized reward $\mathbf{\tilde{r}} \gets \mathbf{r_t} - \lambda \mathbf{c_t}$
    \State Update Q function: $\psi \gets \psi - \eta \nabla_\psi \gL(\psi)$ \Comment{Eq.~\ref{eq:q-loss}}
    \State Update prior, encoder, and high-level policy: $\theta \gets \theta + \eta \nabla_{theta} \gL(\cs)$ \Comment{Eq.~\ref{eq:one-step-obj}}
    \State Update dual variable: $\log \lambda \gets \log \lambda - \eta(C - \E[c_t])$
\EndWhile
\State \textbf{return} $\pi_\theta^z(\ca \mid \cz), \phi_\theta(\cz \mid \cs), m_\theta(\nz \mid \cz, \ca)$
\end{algorithmic}
\end{algorithm}

\subsection{Implementation Details}
We implemented RPC on top of the SAC~\citep{haarnoja2018soft} implementation in TF-Agents~\citep{TFAgents}. Unless otherwise noted, we used the default hyperparameters from that implementation and trained all agents for 3e6 steps. We provide pseudocode in Alg.~\ref{alg:rpc}.

\paragraph{State-based RPC.}
We parameterized the model $m_\theta(\nz \mid \cz, \ca)$ by predicting the \emph{difference} between the $\cz$ and $\nz$. That is, we trained a 2-layer neural network (both layers have 256 units with ReLU activations) to model $p(\nz - \cz \mid \cz, \ca)$. 
We parameterized the encoder $\phi_\theta(\cz \mid \cs)$ as a 2-layer neural network (both layers have 256 units with ReLU activations). Both the model and encoder output the mean and (diagonal) standard deviation of a multivariate Normal distribution. We squashed the mean to be within $[-30, 30]$ and squashed the standard deviation to be within $[0.1, 10.0]$. We used the following function for squashing:
\begin{verbatim}
def squash_to_range(t, low=-np.inf, high=np.inf):
  if low == -np.inf:
    t_low = t
  else:
    t_low = -low * tf.nn.tanh(t / (-low))
  if high == np.inf:
    t_high = t
  else:
    t_high = high * tf.nn.tanh(t / high)
  return tf.where(t < 0, t_low, t_high)
\end{verbatim}
Unless otherwise noted, we set the dimension of $\cz$ to 50, though found this parameter has little effect.

We parameterized the high-level policy $\pi_\theta^z(\ca \mid \cs)$ as a 2-layer neural network (both layers have 256 units with ReLU activations). Following the TF-Agents implementation of SAC, this network first predicts a Normal distribution, which is then squashed by \texttt{tf.tanh} to only output actions within the action space.

\paragraph{Image-based RPC.}
Our image-based version of RPC was based on DrQ~\citep{kostrikov2020image}. Unless otherwise specified, we used the same hyperparameters and architecture as that paper. For example, we used the same action repeat (4 actions) and frame stack (3 frames) as that paper. The only architectural difference was the policy network. We parameterized the encoder $\phi_\theta(\cz \mid \cs)$ as follows:
{\scriptsize
\begin{verbatim}
tf.keras.Sequential([
  tf.keras.layers.Lambda(lambda t: tf.cast(t, tf.float32) / 255.0),  # Normalize image
  tf.keras.layers.Conv2D(filters=32, kernel_size=(3, 3), strides=(2, 2), activation='relu'),
  tf.keras.layers.Conv2D(filters=32, kernel_size=(3, 3), strides=(1, 1), activation='relu'),
  tf.keras.layers.Conv2D(filters=32, kernel_size=(3, 3), strides=(1, 1), activation='relu'),
  tf.keras.layers.Conv2D(filters=32, kernel_size=(3, 3), strides=(1, 1), activation='relu'),
  tf.keras.layers.Flatten(),
  tf.keras.layers.Dense(tfp.layers.IndependentNormal.params_size(latent_dim), activation=_activation),
  tfp.layers.IndependentNormal(latent_dim),
])
\end{verbatim}}
We used the same model and high-level policy as in the state-based experiments. Following DrQ

\paragraph{Dual parameter $\lambda$.}
We updated the dual parameter $\lambda$ via dual gradient ascent to satisfy the bitrate constraint. We parametrized this variable as $\log \lambda$ to ensure that it remains positive. We initialized $\lambda = 1e-6 \approx 0$, and performed updates using \texttt{tf.keras.optimizers.Adam(learning\_rate=3e-4)}. We confirmed via visual inspection that the bitrate constraint was satisfied.

\paragraph{RNN baseline.}
We based the RNN baseline experiments on SAC-RNN implementation in TF-Agents~\citep{TFAgents}.\footnote{\scriptsize \url{https://github.com/tensorflow/agents/blob/master/tf_agents/agents/sac/examples/v2/train_eval_rnn.py}} We trained the agent on sequences of length 20. For fair comparison, we used the same number of \emph{transitions} per batch as other agents that trained on individual transitions. The default batch size in TF-Agent's implementation of SAC is 256, so we trained the RNN on batches with \mbox{256 // 20 = 13} sequences.

\paragraph{Model-based baselines.}
The state-space model and latent-space models used the same model architecture as RPC. This model was trained using standard maximum likelihood with the same optimizer that RPC used for optimizing $m(\nz \mid \cz, \ca)$.

\subsection{Learning Compressed Policies (Fig.~\ref{fig:bits_vs_reward})}
\begin{figure}[H]
    \centering
    \includegraphics[width=\linewidth]{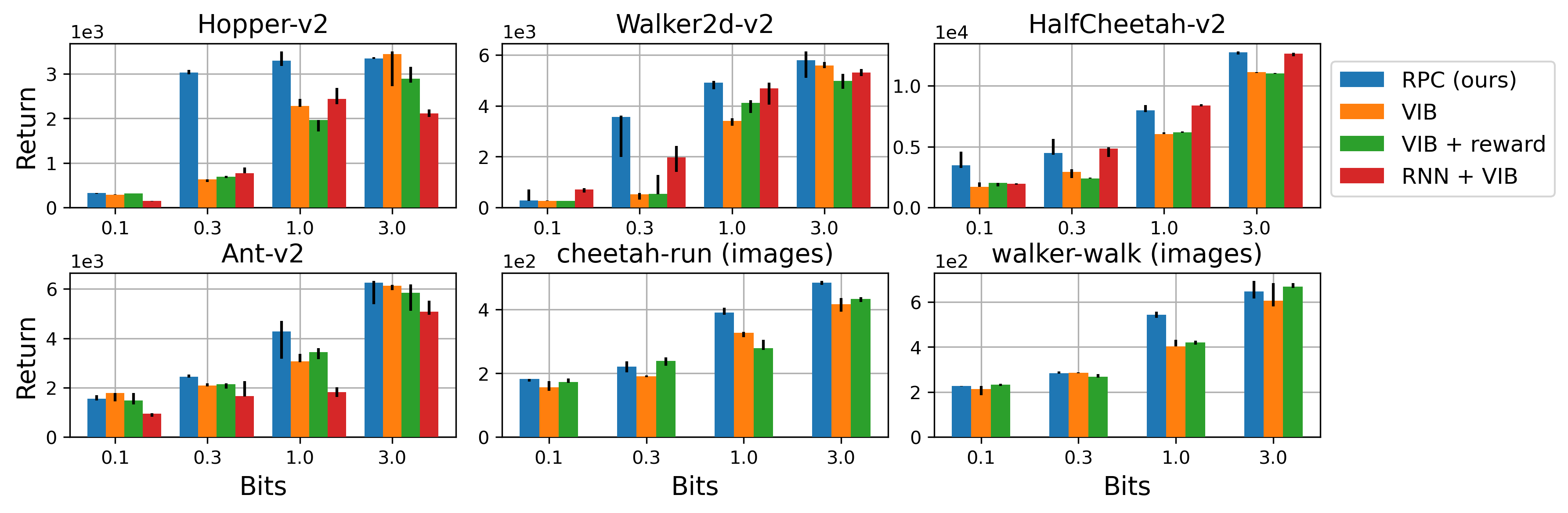}
    \caption{\textbf{Learning Compressed Policies.} This plot shows the same experiment as Fig.~\ref{fig:bits_vs_reward} on a wider range of bitrates, without using normalized returns.}
    \label{fig:bits_vs_reward_appendix}
\end{figure}

We ran this experiment with 5 random seeds, with error bars depicting the [25\%, 75\%] quantiles across random seeds. To compute the return of each policy, we took the average of the last 50 evaluations, corresponding to the last 5e5 training steps (out of a total of 3e6 training steps).
For each task, we normalized the return by the median return of the best performing baseline. Fig.~\ref{fig:bits_vs_reward_appendix} plots returns without normalization.

\subsection{Behavior of Compressed Policies (Fig.~\ref{fig:two_way_qualitative} (bottom))}
We used the \texttt{two-way} task from~\citet{highwayenv}. We initially observed that the agent often crashed into other cars, so we modified the reward function by adding a penalty of -50 when the agent crashes. The original environment has discrete actions, so we added a wrapper around the environment that allows RPC to command continuous actions, which would then be discretized. We trained RPC for 1e5 steps on this environment.

\subsection{Behavior of Compressed Policies (Fig.~\ref{fig:acc_qualitative})}
For this experiment, we used a modified version of the \texttt{two-way} task from~\citet{highwayenv}. We modified the environment to disable passing, so the car was forced to remain in the same lane. To encourage aggressive driving, we modified the reward function to be $(x / 2000)^{10}$, where $x$ is the distance traveled. We trained RPC for 2e5 steps. The ``Compressed Policy'' refers to RPC using a bitrate constraint of 0.1.

\subsection{Value of Information (Fig.~\ref{fig:representations} (top))}
\label{sec:voi-experiment}
The value of observing $\cz$ is given by
\begin{equation*}
    \text{ValueOfInfo}(\nz) = \E \left[\sum_t \gamma^t r(\cs, \ca) \mid \nz \right] - \E \left[\sum_t \gamma^t r(\cs, \ca) \mid \cz \right],
\end{equation*}
while the cost of observing $\cz$ is given by
\begin{equation*}
    \text{CostOfInfo}(\nz) = \E[\log \phi(\nz \mid \ns) - \log m(\nz \mid \cz, \ca)].
\end{equation*}
Intuitively, the optimal policy will only choose to look at representations $\cz$ when $\text{ValueOfInfo}(\nz) > \text{CostOfInfo}(\nz)$. We compare these two quantities in the experiments (Fig.~\ref{fig:representations} (top)). Note that both the cost of information and value of information depend on the encoder,~$\phi(\cs)$.

For this experiment, we applied RPC to the \texttt{HalfCheetah-v2} environment using a bitrate constraint of~10.0.

\subsection{Sparse Representations (Fig.~\ref{fig:representations} (bottom))}

For this experiment, we applied RPC to the \texttt{HalfCheetah-v2} environment. Since both the encoder $\phi(\cz \mid \cs)$ and model $m(\nz \mid \cz, \ca)$ predict \emph{diagonal} normal distributions, we can compute the KL divergence between each coordinate. For the plot in Fig.~\ref{fig:representations} (bottom), we sorted these KLs before plotting.

\subsection{Hierarchical RL (Fig.~\ref{fig:hrl})}

\paragraph{Pre-training.}
We learned the representation of actions by applying RPC to the \texttt{push-v2} task in Metaworld~\citep{yu2020meta}. To make the subsequent optimization easier, we set the dimension of $\cz$ to 3.

\begin{wrapfigure}{R}{0.5\textwidth}
\begin{minipage}{0.5\textwidth}
\begin{minted}{python}
def objective_fn(z_list, horizon):
  env.reset()
  total_r = 0.0
  for z in z_list:
    for _ in range(horizon):
      a = policy(z)
      _, r, done, _ = env.step(a)
      total_r += r
      if done:
        break
      z = prior(z, a)
  loss = -1 * total_r
cma.fmin(objective_fn)
\end{minted}
\end{minipage}
\caption{Pseudocode for hierarchical RL} \label{fig:pseudocode}
\end{wrapfigure}
\paragraph{Hierarchical RL.}
We optimize over the list of $\cz$s using CMA-ES~\citep{hansen2006cma}. We provide pseudocode in Fig.~\ref{fig:pseudocode}. The aim of this approach is to test whether compression results in behaviors that can transfer to new tasks. We use CMA-ES as a black-box optimizer because of its simplicity, and emphasize that more sophisticated optimization routines (e.g., actor-critic RL with actions corresponding to $\cz$) would likely perform better.

For the \texttt{pusher} task (which was the same as the training task), we used one behavior $\cz$ for a horizon of 100 steps. For the \texttt{pusher\_wall} task, we used two behaviors $\cz$, each with horizon 100 steps. For the \texttt{button} task, we used one behavior $\cz$ for a horizon of 100 steps. For the \texttt{drawer\_open} task, we used two behaviors $\cz$, each with a horizon of 75 steps.

\paragraph{Action Repeat.}
In some tasks, actions correspond to a desired pose. Thus, action repeat would correspond to moving to a desired pose, which seemed like a reasonable baseline in these manipulation tasks. Our experiments highlight that the behaviors learned by RPC do more than move to a particular pose. In addition, the behaviors learned by RPC do obstacle avoidance and objective manipulation.

\subsection{Robustness to missing observations (Fig.~\ref{fig:sensor_dropout})}
For this experiment, we trained RPC using a bitrate of 0.1 of \texttt{HalfCheetah-v2} and 3.0 for \texttt{Walker2d-v2}.
We evaluated all methods by dropping observations independently. Note that none of the methods were trained using observation dropout, so this experiment explicitly tests robustness to new disturbances introduced at test time.

\subsection{Adversarial Robustness to Dynamics (Fig.~\ref{fig:pgd} (left))}
The adversary aims to apply a small perturbation to that state to make the policy perform as poorly as possible. The adversary's objective is 
\begin{equation*}
    \min_{\mathbf{s_\text{adv}} \in \gB(\mathbf{s})} V^\pi(\mathbf{s_\text{adv}}) = Q^\pi(\mathbf{s_\text{adv}}, \mathbf{a} \sim \pi(\mathbf{a} \mid \mathbf{s_\text{adv}})).
\end{equation*}
We instantiate the adversary using projected gradient descent~\citep{madry2017towards} with step size 0.1.
For fair comparison, we evaluate each policy on states collected by rolling out that policy.
This experiment used the \texttt{Ant-v2} environment, applying RPC using a bitrate of 0.3. We repeated this attack on 20 states sampled from each policy's state distribution. The dark line shows the average across these 20 attacks. As expected, the agent is more vulnerable to attacks in some states than in other states.

\subsection{Adversarial Robustness to Observations (Fig.~\ref{fig:pgd} (right))}
We optimize the adversary to corrupt the state using the following objective:
\begin{equation*}
    \min_{\mathbf{s_\text{adv}} \in \gB(\mathbf{s})} Q^\pi(\mathbf{s}, \mathbf{a} \sim \pi(\mathbf{a} \mid \mathbf{s_\text{adv}})).
\end{equation*}
The adversary is optimized using PGD, using three steps of size 0.1 for each state.
Note that we only change the observation, not the true state of the environment. This difference from the previous experiment is important, as it allows us to unroll the policy and adversary for an entire trajectory.
We used the \texttt{Ant-v2} environment for this experiment.

\subsection{Robust RL (Fig.~\ref{fig:robust-rl})}

We followed prior work in setting up the robust RL experiments~\citep{tessler2019action}.
To modify the mass, we scaled the \texttt{env.model.body\_mass} attribute of the environment by a fixed constant. To modify friction, we scaled the \texttt{env.model.geom\_friction} attribute of the environment by a fixed constant. We used perturbations much larger than prior work~\citep{tessler2019action} because we found that standard RL was already robust to smaller ranges of perturbations.

\section{Proofs}
\label{appendix:proofs}

\subsection{Model Compression is a Lower Bound on Open-Loop Control (Lemma~\ref{lemma:open-loop})}

In this section we formally define the assumptions for Lemma~\ref{lemma:open-loop} and then provide the proof.

First, we define an open loop policy with hidden state $\cz$. This policy updates its hidden state as \mbox{$m(\nz \mid \cz, \ca)$} and produces actions by sampling $p(\ca \mid \cz)$.
Note that the open-loop sequence of actions generated by the policy is evaluated under the \emph{true} system dynamics, $p(\ns \mid \cs, \ca)$. The open loop policy does not observe the transitions from the true system dynamics.

The main idea of the proof will be to apply an evidence lower bound on the expected reward objective. This idea alone almost completes the proof. The main technical challenge is accounting for discount factors: a na\"ive application of an ELBO will result in discounted rewards but an undiscounted information term.

\begin{proof}
The \textbf{first step} is to recognize that the expected discounted return objective can be written as the expected \emph{terminal} reward of a mixture of finite-length episodes. Define $p_H(\tau)$ as a distribution over length-$H$ episodes. We can then write the expected discounted return objective as follows:
\begin{align*}
    \E_{p(\mathbf{\tau})}\left[\sum_{t=1}^\infty \gamma^t r(\cs, \ca) \right]
    &= \sum_{H=1}^\infty \gamma^H \E_{p_H(\mathbf{\tau})}\left[r(\mathbf{s_H}, \mathbf{a_H}) \right].
\end{align*}
We can now obtain a lower bound on the \emph{log} of the expected return objective:
\begin{align}
    \log & \E_{p(\mathbf{s_{1:\infty}}, \mathbf{a_{1:\infty}}, \mathbf{z_{1:\infty}})}\left[\sum_{t=1}^\infty \gamma^t r(\cs, \ca) \right] \nonumber \\
    &= \log \left( \sum_{H=1}^\infty \gamma^H \E_{p(\mathbf{s_{1:H}}, \mathbf{a_{1:H}}, \mathbf{z_{1:H}})}\left[r(\mathbf{s_H}, \mathbf{a_H}) \right] \right) \nonumber \\
    &= \log \left( \frac{1 - \gamma}{\gamma} \sum_{H=1}^\infty \gamma^H \E_{p(\mathbf{s_{1:H}}, \mathbf{a_{1:H}}, \mathbf{z_{1:H}})}\left[r(\mathbf{s_H}, \mathbf{a_H}) \right] \right) + \log \frac{\gamma}{1 - \gamma} \nonumber \\
    &\ge \frac{(1 - \gamma)}{\gamma} \sum_{H=1}^\infty \gamma^H \log \left(\E_{\mathbf{p(s_{1:H}}, \mathbf{a_{1:H}}, \mathbf{z_{1:H}})}\left[r(\mathbf{s_H}, \mathbf{a_H}) \right] \right) + \log \frac{\gamma}{1 - \gamma}. \label{eq:step-1}
\end{align}
The inequality is Jensen's inequality.
The constant $\frac{1 - \gamma}{\gamma}$ was introduced because Jensen's inequality must be applied to a proper probability distribution.

The \textbf{second step} is to define a variational distribution $q(\mathbf{s_{1:\infty}}, \mathbf{a_{1:\infty}}, \mathbf{z_{1:\infty}})$ as
\begin{equation*}
    q(\mathbf{s_{1:\infty}}, \mathbf{a_{1:\infty}}, \mathbf{z_{1:\infty}}) = \prod_t p(\ca \mid \cz) q(\cz \mid \cs) p(\ns \mid \cs, \ca).
\end{equation*}
We can then obtain a lower bound on Eq.~\ref{eq:step-1}
{\footnotesize
\begin{align}
    \log & \E_{p(\mathbf{s_{1:\infty}}, \mathbf{a_{1:\infty}}, \mathbf{z_{1:\infty}})}\left[\sum_{t=1}^\infty \gamma^t r(\cs, \ca) \right] \nonumber \\
    & \ge \frac{1 - \gamma}{\gamma} \sum_{H=1}^\infty \gamma^H \E_{q(\mathbf{s_{1:H}}, \mathbf{a_{1:H}}, \mathbf{z_{1:H}})}\left[\log r(\mathbf{s_H}, \mathbf{a_H}) + \log p(\mathbf{s_{1:H}}, \mathbf{a_{1:H}}, \mathbf{z_{1:H}}) - \log q(\mathbf{s_{1:H}}, \mathbf{a_{1:H}}, \mathbf{z_{1:H}})\right] + \log \frac{\gamma}{1 - \gamma} \nonumber \\
    & = \frac{1 - \gamma}{\gamma} \sum_{H=1}^\infty \gamma^H \E_{q(\mathbf{s_{1:H}}, \mathbf{a_{1:H}}, \mathbf{z_{1:H}})}\left[\log r(\mathbf{s_H}, \mathbf{a_H}) + \sum_{t=1}^H \log m(\nz \mid \cz, \ca) - \log \phi(\cz \mid \cs)\right] + \log \frac{\gamma}{1 - \gamma}. \label{eq:step-2}
\end{align}}
The final line follows from simplifying the definitions of $p(\mathbf{s_{1:\infty}}, \mathbf{a_{1:\infty}}, \mathbf{z_{1:\infty}})$ and $q(\mathbf{s_{1:\infty}}, \mathbf{a_{1:\infty}}, \mathbf{z_{1:\infty}})$. Note that we have used the assumption that $r(\cs, \ca) > 0$ to ensure that $\log r(\cs, \ca)$ is well defined.

Our \textbf{third step} is to do the opposite of the first step:
recognize that expectations over a mixture of finite-horizon episodes are equivalent to expectations over a discounted infinite-length episode. With this, we can rewrite Eq.~\ref{eq:step-2} as follows:
{\footnotesize
\begin{align}
    \log & \E_{p(\mathbf{s_{1:\infty}}, \mathbf{a_{1:\infty}}, \mathbf{z_{1:\infty}})}\left[\sum_{t=1}^\infty \gamma^t r(\cs, \ca) \right] \nonumber \\
    & \ge \frac{1 - \gamma}{\gamma} \E_{q(\mathbf{s_{1:\infty}}, \mathbf{a_{1:\infty}}, \mathbf{z_{1:\infty}})}\left[ \sum_{H=1}^\infty \gamma^H \left( \log r(\mathbf{s_H}, \mathbf{a_H}) + \sum_{t=1}^H \log m(\nz \mid \cz, \ca) - \log \phi(\cz \mid \cs) \right) \right] + \log \frac{\gamma}{1 - \gamma}. \label{eq:step-3}
\end{align}}
We can simplify the double summation using the following identity, where $\mathbf{x_t}$ is shorthand for $\log m(\nz \mid \cz, \ca) - \log \phi(\cz \mid \cs)$:
\begin{align*}
    \sum_{H=1}^\infty \gamma^H \sum_{t=1}^H \mathbf{x_t}
    &= \gamma (\mathbf{x_1}) + \gamma^2 (\mathbf{x_1} + \mathbf{x_2}) + \gamma^3 (\mathbf{x_1} + \mathbf{x_2} + \mathbf{x_3}) + \cdots \\
    &= \mathbf{x_1} (\gamma + \gamma^2 + \gamma^3 + \cdots) + \mathbf{x_2} (\gamma^2 + \gamma^3 + \cdots) + \cdots \\
    &= \mathbf{x_1} \frac{\gamma}{1 - \gamma} + \mathbf{x_2} \frac{\gamma^2}{1 - \gamma} + \mathbf{x_3} \frac{\gamma^3}{1 - \gamma} + \cdots \\
    &= \frac{1}{1 - \gamma} \sum_{t=1}^\infty \gamma^t \mathbf{x_t}.
\end{align*}
Applying this identity to Eq.~\ref{eq:step-3}, we get
{\footnotesize
\begin{align*}
    \log & \E_{p(\mathbf{s_{1:\infty}}, \mathbf{a_{1:\infty}}, \mathbf{z_{1:\infty}})}\left[\sum_{t=1}^\infty \gamma^t r(\cs, \ca) \right] \nonumber \\
    & \ge \frac{1 - \gamma}{\gamma} \E_{q(\mathbf{s_{1:\infty}}, \mathbf{a_{1:\infty}}, \mathbf{z_{1:\infty}})}\left[ \sum_{t=1}^\infty \gamma^t \left( \log r(\cs, \ca) + \frac{1}{1 - \gamma}\log m(\nz \mid \cz, \ca) - \frac{1}{1 - \gamma} \log \phi(\cz \mid \cs) \right) \right] + \log \frac{\gamma}{1 - \gamma} \\
    & \ge \frac{1}{\gamma}\E_{q(\mathbf{s_{1:\infty}}, \mathbf{a_{1:\infty}}, \mathbf{z_{1:\infty}})}\left[ \sum_{t=1}^\infty \gamma^t \left( (1 - \gamma) \log r(\cs, \ca) + \log m(\nz \mid \cz, \ca) - \log \phi(\cz \mid \cs) \right) \right] + \log \frac{\gamma}{1 - \gamma}.
\end{align*}}
Exponentiating both sides, we obtain the desired result.

\end{proof}
This derivation is useful because it relates model compression to open-loop control. Precisely, it says that a compressed model is one that will perform well under open loop rollouts.

\subsection{Proof of Lemma~\ref{lemma:open-loop-bound}}

\begin{proof}
To start, we define the distribution over trajectories $\mathbf{\tau} = (\mathbf{s_{1:\infty}}, \mathbf{a_{1:\infty}}, \mathbf{z_{1:\infty}})$ produced by $\pi^\text{open}$ and $\pi^\text{reactive}$:
\begin{align*}
    p^\text{open}(\mathbf{\tau}) &= p_1(\mathbf{s_1}) \prod_t p(\ns \mid \cs, \ca) \pi^z(\ca \mid \cz) m(\nz \mid \cz, \ca) \\
    p^\text{reactive}(\mathbf{\tau}) &= p_1(\mathbf{s_1}) \prod_t p(\ns \mid \cs, \ca) \pi^z(\ca \mid \cz) \phi(\cz \mid \cs).
\end{align*}
Note that $R(\mathbf{\tau}) \triangleq \sum_t \gamma^t r(\cs, \ca)$ is a deterministic function of a random variable, $\mathbf{\tau}$.
We can now write the difference in returns as $\E_{\pi^\text{reactive}}[R(\mathbf{\tau})] - \E_{\pi^\text{open}}[R(\mathbf{\tau})]$. The main idea of our proof will be to show that this difference is not too large.
\begin{align}
    \E_{\pi^\text{reactive}}[R(\mathbf{\tau})] - \E_{\pi^\text{open}}[R(\mathbf{\tau})]
    &= \int R(\mathbf{\tau}) (p^\text{reactive}(\mathbf{\tau}) - p^\text{open}(\mathbf{\tau})) d \mathbf{\tau} \nonumber \\
    & \stackrel{(a)}{\le} R_\text{max} \int |(p^\text{reactive}(\mathbf{\tau}) - p^\text{open}(\mathbf{\tau})) | d \mathbf{\tau} \nonumber \\
    & \stackrel{(b)}{\le} R_\text{max} \sqrt{\frac{1}{2} KL(p^\text{reactive}(\mathbf{\tau})\| p^\text{open}(\mathbf{\tau}))}. \label{eq:return-diff}
\end{align}
We used H\"older's inequality in \emph{(a)} and Pinsker's inequality in \emph{(b)}. Our second step is to relate the KL divergence to the compression objective.
\begin{align*}
    KL(&p^\text{reactive}(\mathbf{\tau}) \| p^\text{open}(\mathbf{\tau}))\\
    &= \E_{p^\text{reactive}}\left[\cancel{\log p_1(\mathbf{s_1})} + \sum_t \cancel{p(\ns \mid \cs, \ca)} + \cancel{\log \pi^z(\ca \mid \cz)} + \log \phi(\cz \mid \cs) \right. \\
    & \qquad\qquad \left. - \cancel{\log p_1(\mathbf{s_1})} - \sum_t \left( \cancel{p(\ns \mid \cs, \ca)} + \cancel{\log \pi^z(\ca \mid \cz)} + \log m(\nz \mid \cz, \ca)) \right) \right] \\
    &= \E_{p^\text{reactive}}\left[\sum_t \log \phi(\cz \mid \cs) - \log m(\nz \mid \cz, \ca) \right] \le C.
\end{align*}
where $C$ is the information constraint (Eq.~\ref{eq:objective}).
We then substitute this simplified expression for the KL into Eq.~\ref{eq:return-diff}:
\begin{equation*}
    \E_{\pi^\text{reactive}}[R(\mathbf{\tau})] - \E_{\pi^\text{open}}[R(\mathbf{\tau})] \le R_\text{max}\sqrt{\frac{1}{2} C}.
\end{equation*}
Rearranging terms, we obtain the desired result:
\begin{equation*}
    \E_{\pi^\text{open}}[R(\mathbf{\tau})] \ge \E_{\pi^\text{reactive}}[R(\mathbf{\tau})] - R_\text{max}\sqrt{\frac{1}{2} C}.
\end{equation*}
\end{proof}
For this result to be non-vacuous, we need that $\sqrt{\frac{1}{2}C} < 1$, so the average per-transition $c$ must satisfy $c = (1 - \gamma) C \le 2 \cdot (1 - \gamma)$:
\begin{equation*}
    \E_\pi[\log \phi(\cz \mid \cs) - \log m(\nz \mid \cz, \ca)] \le 2 \cdot (1 - \gamma).
\end{equation*}
For example, when $\gamma = 0.99$, we would need that the average per-transition KL be less than 0.02.

\section{Additional Experiments}
\label{appendix:more-experiments}

\begin{figure}[H]
    \centering
    \includegraphics[width=0.8\linewidth]{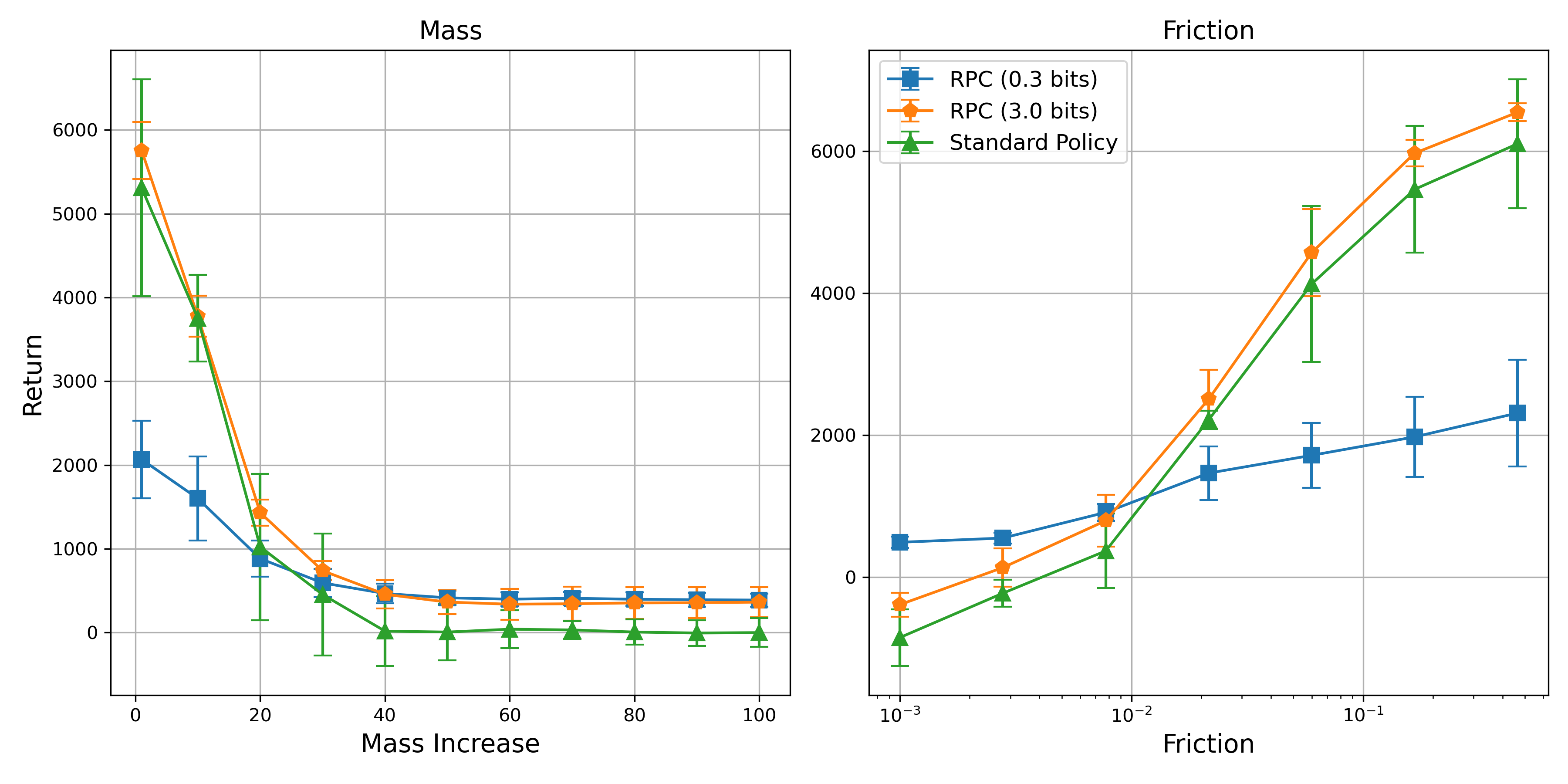}
    \caption{\textbf{Robust RL}: This plot shows the same experiment as Fig.~\ref{fig:robust-rl} with error bars corresponding to the standard deviation across five \emph{training} random seeds.}
    \label{fig:robust-rl-appendix}
\end{figure}

\begin{figure}[H]
    \centering
    \begin{minipage}{0.45\textwidth}
    \includegraphics[width=\linewidth]{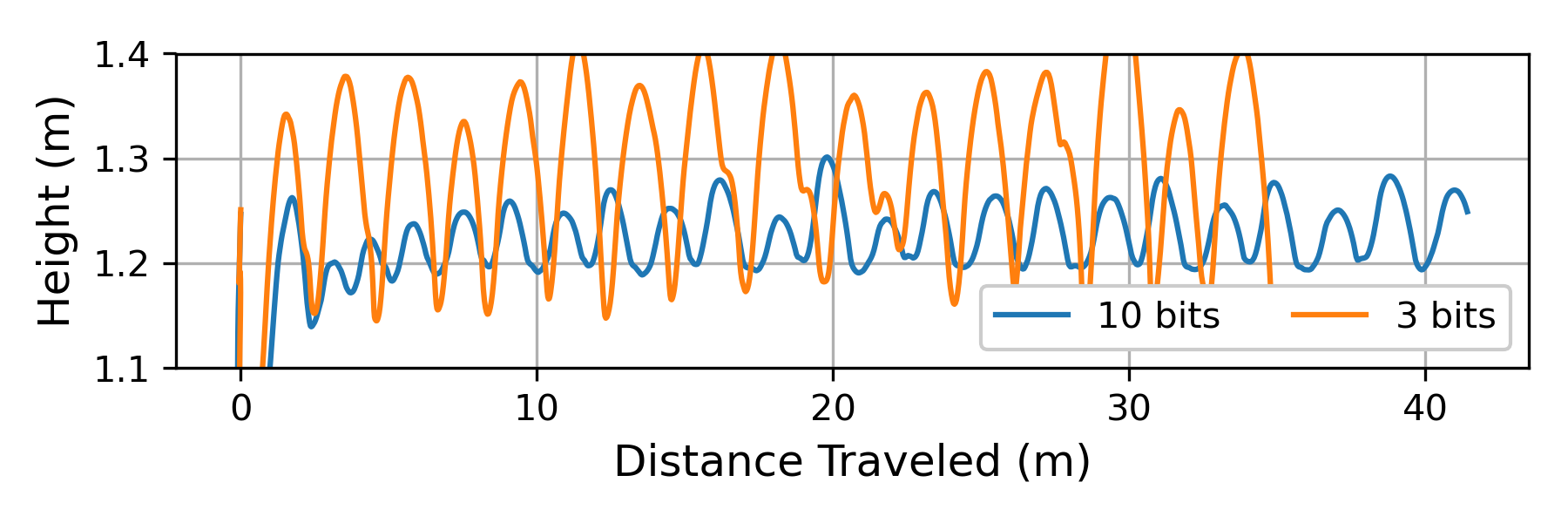}
    \caption{Behavior of compressed walking policies.}
    \label{fig:walker-trace}
    \end{minipage}
    \begin{minipage}{0.45\textwidth}
        \includegraphics[width=\linewidth]{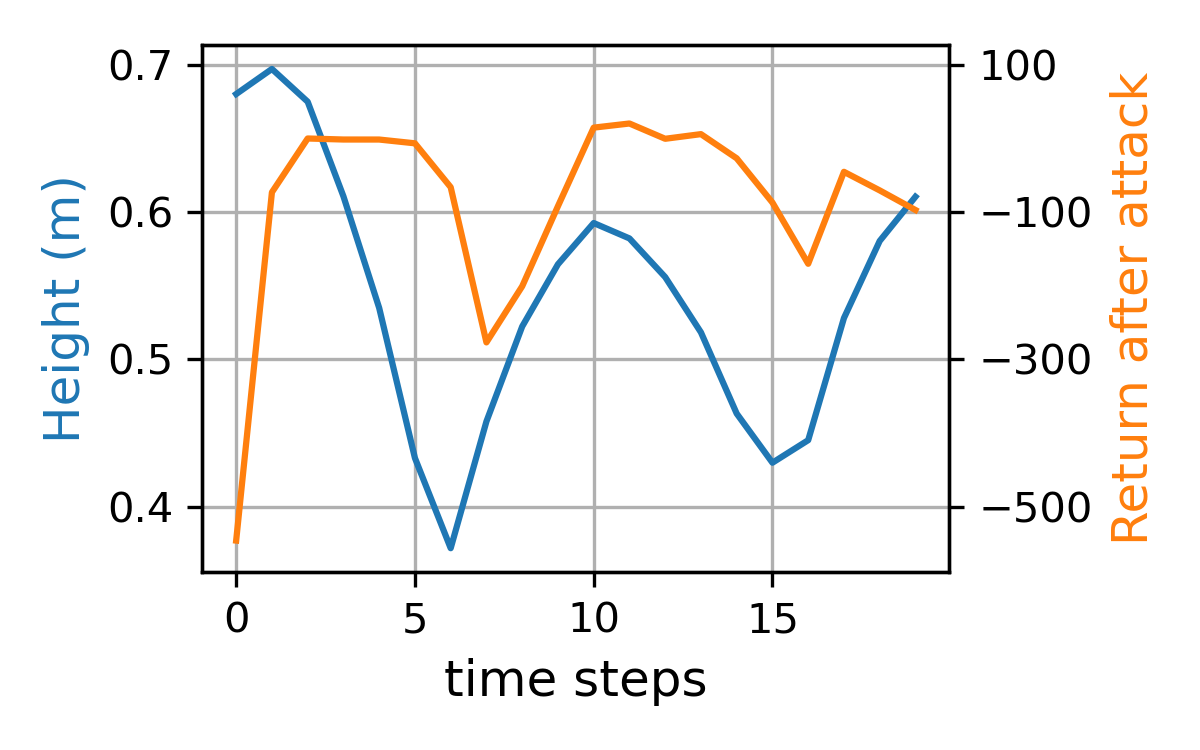}
    \caption{\textbf{Adversarial Robustness}: The agent is most vulnerable just before its feet leave the ground}
    \label{fig:pgd-trace}    
    \end{minipage}
\end{figure}

\paragraph{Visualizing more compressed policies.}
In Fig.~\ref{fig:walker-trace}, we demonstrate that compressed policies also learn different walking gaits on a locomotion task. The agent requires many more bits each time it touches the ground, so the compressed policy takes bigger steps to minimize the number of footsteps.
See the project website for videos comparing the behaviors learned for varying levels of compression.

\paragraph{Behavior of Attacked Policies}
\label{appendix:pgd-trace}

We visualize the behavior of the adversarial attack on dynamics described in Sec.~\ref{sec:experiments-robust}. Visualizing the uncompressed policy in Fig.~\ref{fig:pgd-trace}, we observe that the policy is most vulnerable to attack the moment the robot launches off the ground for each step. This makes sense, as the actions right before takeoff dictate the path taken when the robot is in the air before the next step.

\end{document}